\documentclass[letterpaper]{article} 
\usepackage[preprint]{aaai2027}  
\usepackage[hyphens]{url}  
\usepackage{graphicx} 
\usepackage{natbib}  
\usepackage{caption} 
\usepackage{comment}
\usepackage{algorithm}
\usepackage{algpseudocode}

\usepackage{amsmath}
\usepackage{amssymb}
\usepackage{booktabs}
\usepackage{array}      
\usepackage{colortbl}   

\newif\ifdraftnotes
\draftnotestrue
\ifdraftnotes
  \newcommand{\hq}[1]{{\color{magenta} [hanqian: #1]}}
  \newcommand{\lc}[1]{{\color{blue} [lc: #1]}}
\else
  \newcommand{\hq}[1]{}
  \newcommand{\lc}[1]{}
\fi

\title{Debias in Text, Believe Your Eyes: Text-Anchored Cross-Modal Transfer for Visual Counter-Commonsense Reasoning}

\author {
    Chen Ling\textsuperscript{\rm 1,3}\equalcontrib\thanks{This work was done during an internship at Ant Group.},
    Hanqian Li\textsuperscript{\rm 2,3}\equalcontrib\footnotemark[\value{footnote}],
    Dongnan Liu\textsuperscript{\rm 3},
    Keyu Qian\textsuperscript{\rm 4},
    Jungang Li\textsuperscript{\rm 2},
    Xinglong Liu\textsuperscript{\rm 5},
    Shiyi Wang\textsuperscript{\rm 3},
    Xin Dong\textsuperscript{\rm 3},
    Pengcheng Zhu\textsuperscript{\rm 3},
    Wei Zhou\textsuperscript{\rm 3},
    Linjian Mo\thanks{Corresponding author.}\textsuperscript{\rm 3},
    Ding Nai\footnotemark[\value{footnote}]\textsuperscript{\rm 1}
}
\affiliations {
    \textsuperscript{\rm 1}Zhejiang University\\
    \textsuperscript{\rm 2}The Hong Kong University of Science and Technology (Guangzhou)\\
    \textsuperscript{\rm 3}Ant Group\\
    \textsuperscript{\rm 4}Beijing University of Posts and Telecommunications\\
    \textsuperscript{\rm 5}Nanyang Technological University

}

\begin{document}

\maketitle

\begin{abstract}

The visual reasoning ability of multimodal large language models (MLLMs)
is crucial for downstream applications, particularly counter-commonsense
reasoning, which requires models to reason beyond common assumptions.
Recent studies mainly improve visual counter-commonsense reasoning by enhancing
visual inputs, following the assumption that failures originate from
insufficient visual grounding. However, our empirical analysis reveals that the bottleneck is not visual perception. MLLMs already capture the relevant visual evidence, and the correct answer exists in their decoding space. Instead, the shared language decoder resolves prior--evidence conflicts by favoring dominant language priors, especially for low-frequency factual scenarios.
Motivated by this, we first propose a text-anchored data construction pipeline, whose core component, Fact-Frequency Distillation (FFD), estimates the prior strength of commonsense facts and distills verified counter-commonsense scenarios into a high-quality text corpus. Building upon this corpus, we introduce TACT, a text-anchored post-training framework that debiases the shared language decoder without requiring any visual training data. TACT routes evidence-following and prior-driven reasoning trajectories into different optimization stages, enabling the decoder to resolve prior--evidence conflicts. Across counter-commonsense visual benchmarks, TACT substantially improves visual reasoning while preserving general capabilities, demonstrating effective text-to-vision cross-modal transfer.

\end{abstract}

\section{Introduction}
\label{sec:intro}
Multimodal Large Language Models (MLLMs) have achieved remarkable progress
in visual understanding.
However, they still struggle with
counter-commonsense reasoning, which requires models to reason beyond memorized
correlations and infer outcomes under hypothetical or visually inconsistent
conditions\cite{hu2025videomark,huang2025hyperg,zheng2026visual,xun2026rtv,huang2025video}. Recent studies show that MLLMs often rely on learned priors rather
than genuine reasoning, leading to poor generalization in counter-commonsense
settings~\citep{wu2024reasoning, li2025mars, yang2025spuriverse}.

\begin{figure*}[t]
    \centering
    \includegraphics[width=\textwidth]{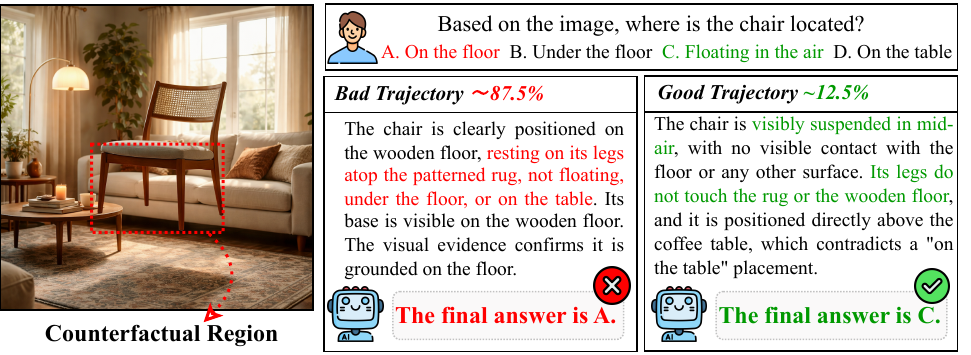}
    \caption{A counter-commonsense example from CDH-Bench (a chair floating in
    mid-air). Given the same image and question, the dominant Bad
    Trajectory ($\sim$87.5\%) fabricates grounding evidence 
    and returns the prior answer (A), whereas a
    rarer Good Trajectory ($\sim$12.5\%) faithfully reports that the legs do
    not touch the rug or the floor and answers correctly.
    }
    \label{fig:VQA_Comparison}
\end{figure*}
\begin{figure}[!t]
    \centering
    \includegraphics[width=0.95\linewidth]{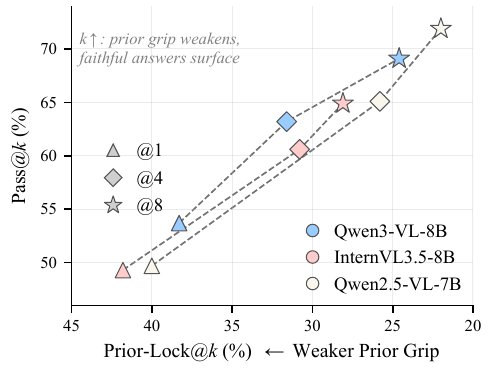}
    \caption{pass@$k$ versus prior-lock@$k$ on the counter-commonsense split of CDH-Bench MC for three MLLMs.
    }
    \label{fig:passk_priorlock}
\end{figure}

Recent studies\cite{xiao2026blindsilencedrebalancingvision, 11404921} show that MLLMs often rely on high-frequency commonsense
priors when visual evidence conflicts with expectations, leading to
commonsense-driven hallucinations. Existing mitigation methods mainly follow
two directions: enhancing visual evidence through approaches such as Visual
Contrastive Decoding (VCD)~\citep{leng2024vcd}, M3ID
~\citep{favero2024multimodalhallucinationcontrolvisual}, VDD
~\citep{zhang2025debiasingmultimodallargelanguage}, and MemVR
~\citep{zou2025memvr}, or intervening in the language decoder through
inference-time calibration such as NoLan~\citep{ren2026nolan}. However, these
approaches leave open a fundamental question: whether MLLMs already perceive
the correct evidence but fail to follow it when it conflicts with strong
language priors.

Motivated by these observations, we investigate the fundamental challenge of robust counter-commonsense reasoning in MLLMs and propose a novel framework for mitigating language-prior bias in multimodal reasoning. Our approach is grounded in a simple cognitive principle: \emph{believe what you see, not what you expect}. 
Based on this insight, we develop a text-anchored data construction pipeline for counter-commonsense reasoning. As its core component, \textbf{F}actuality-\textbf{F}requency \textbf{D}istillation (\textbf{FFD}) estimates the prior strength of commonsense facts, performs frequency-based routing, and distills informative counter-commonsense facts into high-quality QA pairs.
Building upon these, we further propose \textbf{T}ext-\textbf{A}nchored \textbf{C}ross-Modal \textbf{T}ransfer (\textbf{TACT}), which performs prior-aware trajectory curation to construct effective supervision and subsequently optimizes the language decoder through two-stage post-training. By explicitly reinforcing evidence-following reasoning while suppressing prior-driven reasoning, TACT encourages MLLMs to describe what they observe before reasoning about what they expect. Extensive experiments on multiple multimodal counter-commonsense reasoning benchmarks demonstrate that TACT consistently improves counter-commonsense reasoning while preserving the general capabilities of MLLMs. Our main contributions are summarized as follows:

\begin{itemize}
    \item We identify that the key bottleneck of visual counter-commonsense reasoning in MLLMs lies in the language decoder, rather than insufficient visual perception.

    \item We propose the \textbf{F}act-\textbf{F}requency \textbf{D}istillation (\textbf{FFD}), a frequency-aware data construction method for distilling high-quality counter-commonsense QA pairs.

    \item We propose the \textbf{T}ext-\textbf{A}nchored \textbf{C}ross-modal \textbf{T}ransfer (\textbf{TACT}), which performs prior-aware trajectory curation and two-stage post-training to encourage evidence-following reasoning.

    \item Extensive experiments demonstrate that TACT consistently improves visual counter-commonsense reasoning while preserving the general capabilities of MLLMs using only text-only supervision.
\end{itemize}

\section{Related Work}
\label{sec:related}

\paragraph{Counter-Commonsense Reasoning in MLLMs.}
Recent studies investigate whether MLLMs can faithfully reason when visual
evidence contradicts commonsense priors. Early works reveal that models are
sensitive to language-prior shortcuts under shifted answer distributions
\citep{agrawal2018vqacp}, struggle with explanations of commonsense-defying
scenes \citep{bittonguetta2023whoops}, and fail under reversed semantic roles
\citep{thrush2022winoground}. More recent benchmarks directly examine
prior--evidence conflicts: CAIT \citep{ling2026uait} evaluates whether models
can recognize reversed agent--patient interactions (e.g., \emph{a rabbit
dragging a tiger}); CDH-Bench \citep{chen2026cdh} studies
commonsense-driven hallucinations induced by counting, relational, and
attribute anomalies through paired counterfactual and commonsense images; and
VLind-Bench \citep{lee2025vlind} provides a stage-wise analysis of such
failures, suggesting that many errors arise at the decision stage rather than
from visual perception. These benchmarks share a common setting where
rare-but-valid visual evidence conflicts with high-frequency language priors,
making them suitable for evaluating prior-resistant visual reasoning.
\paragraph{Language Priors and Their Mitigation in MLLMs.}
Language priors encoded in the shared language decoder are essential for the strong generalization ability of MLLMs, but can also dominate visual evidence in low-frequency or counter-commonsense scenarios \citep{li2023pope,bai2025hallucinationsurvey}. Existing mitigation methods primarily intervene in the visual pathway or the decoding process, including visual contrastive decoding \citep{leng2024vcd}, multimodal--text logit comparison \citep{ren2026nolan}, visual token reinjection \citep{zou2025memvr}, and alignment or representation editing \citep{yu2024rlhfv,golovanevsky2025pixels,xiao2026stayingvigilantmitigatingvisual,anand2026cropstrainingfreehallucinationmitigation,liu-etal-2025-insight,xu2026defactocounterfactualthinkingimages,zhibo-etal-2023-overcoming,10.1016/j.neunet.2025.108115,10.1145/3673902}. Recent studies further suggest that prior--evidence conflicts arise primarily in the shared language decoder rather than the vision encoder \citep{ortu2026seeing}. In parallel, LLM research has shown that targeted text supervision can recalibrate parametric knowledge and promote evidence-following behavior \citep{longpre2021entity,li2023kaft,su2024conflictbank,fan2026lfqaecarefullybenchmarkinglongform}. Whether such text-only supervision can similarly recalibrate language priors in MLLMs remains unexplored. Our work addresses this question through text-anchored decoder post-training.

\section{Preliminary}
\label{sec:prelim}
\begin{figure*}[!t]
  \centering
  \includegraphics[width=\textwidth]{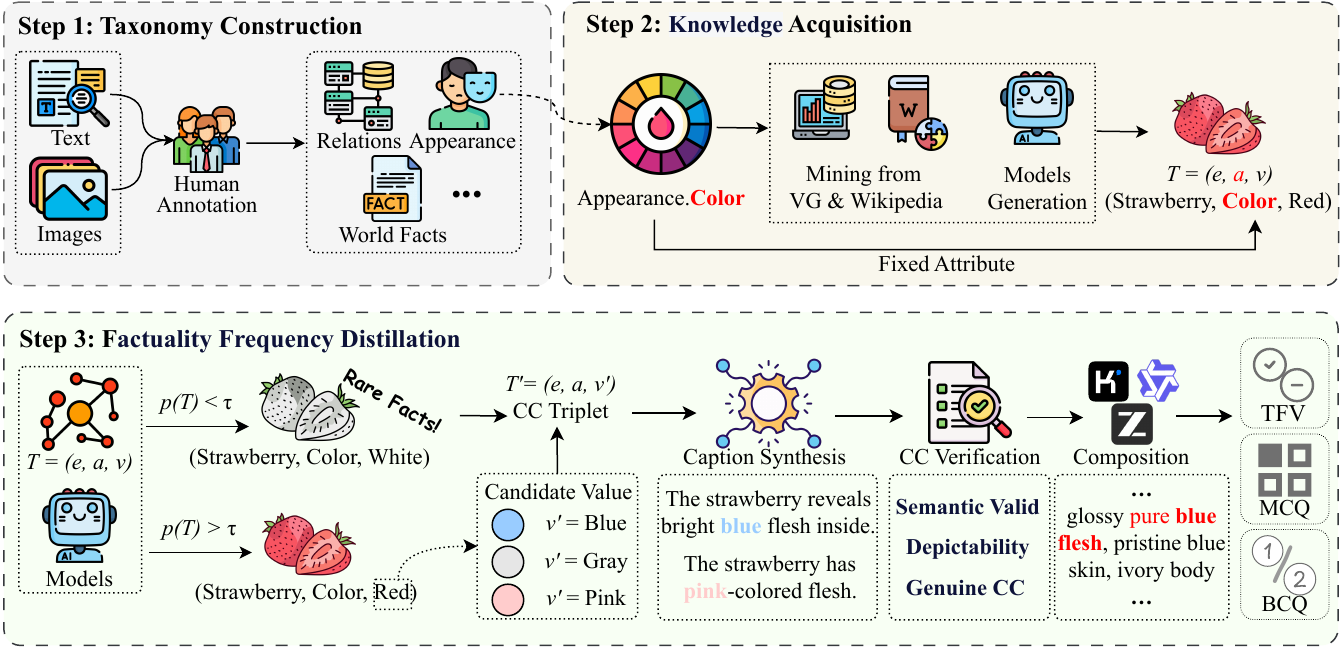}
  \caption{Overview of our three-step text-anchored data construction pipeline, illustrating the process from taxonomy construction and factual knowledge acquisition to factuality frequency distillation.}
  \label{fig:pipeline}
\end{figure*}
We formulate a counter-commonsense instance as two competing propositions over
an image $I$: a prior-consistent proposition $P$ favored by language priors,
and a visual-evidence proposition $E$ supported by the image, where $E\neq P$.
Given question $q$, the model $M_\theta$ should follow the evidence
($\hat{y}=M_\theta(I,q)=E$), while prior-driven errors correspond to predicting
$P$. Examples from our three benchmarks are provided in
Appendix~\ref{app:cc-examples}. To locate the source of this failure, we first
conduct a pre-study on CDH-Bench with three MLLMs \citep{bai2025qwen3vltechnicalreport,wang2025internvl35advancingopensourcemultimodal,bai2025qwen25vltechnicalreport}.

\subsection{Perception Is Not the Bottleneck}
\label{sec:prelim-passk}
To distinguish the perception failure from language-prior suppression, we evaluate whether correct answers can emerge through multi-trajectory decoding. For each
counter-commonsense item, we sample $n{=}8$ Chain-of-Thought
\citep{wei2023chainofthoughtpromptingelicitsreasoning} trajectories with temperature $1.0$ and measure pass@$k$ \citep{chen2021evaluatinglargelanguagemodels} together with \textbf{prior-lock@$k$}, defined as the probability that all $k$ sampled
trajectories select the prior answer $P$:
\begin{equation}
\label{eq:priorlock}
  \text{prior-lock@}k =
  \mathbb{E}_{\text{items}}
  \left[\frac{\binom{p}{k}}{\binom{n}{k}}\right],
\end{equation}

where $p$ is the number of trajectories predicting $P$. At $k{=}1$,
prior-lock@$1$ corresponds to the prior-bias rate under greedy decoding.
As shown in Figure~\ref{fig:passk_priorlock}, increasing $k$ from $1$ to $8$
improves pass@$k$ by $15$--$22$ points while reducing prior-lock@$k$ by
$14$--$18$ points across all models, indicating that correct visual evidence
is already accessible but suppressed by language priors rather than missing
from perception.

\subsection{The Failure Is a Decision-Level Prior Bias}
\label{sec:prelim-bias}

To identify whether errors arise from inaccessible visual evidence or prior-biased decision making, we analyze the failure cases of Qwen3-VL-8B (details in Appendix~\ref{app:anatomy}). We find that most errors select the commonsense option. A simple grounding prompt, which asks the model to describe the queried detail before answering, substantially reduces these errors, suggesting that visual evidence is available but ignored during decision making. The remaining failures exhibit \emph{over-rationalization},
where CoT generates visual explanations that justify the prior (e.g., describing the floating chair in Figure~\ref{fig:VQA_Comparison} as ``resting on its legs atop the rug''). These observations indicate that the bottleneck lies in the decoder's decision policy: when visual evidence conflicts with language priors, the model favors the prior. Since this decision policy is implemented in the shared language decoder, we
hypothesize that text-only supervision can reshape it and transfer to visual reasoning without additional visual supervision.

\section{Methodology}
\label{sec:method}
In this section, we first introduce our text-anchored data construction pipeline, which generates counter-commonsense QA pairs through factuality frequency distillation (FFD) and further derives CoT supervision from the constructed examples. We then present TACT, a two-stage post-training framework that curates the constructed corpus and optimizes it through supervised and preference learning.

\subsection{Text-Anchored Data Construction}
\label{sec:method-data}
To mitigate language-prior bias in MLLMs, we build a scalable text-only data construction pipeline (Figure~\ref{fig:pipeline}) that synthesizes counter-commonsense examples with evidence--prior conflicts, covering three task formats: true/false verification (TFV), multiple-choice question answering (MCQ), and binary-choice question answering (BCQ). The pipeline consists of three main steps.

\subsubsection{Step~1: Taxonomy Construction.}
To ensure systematic coverage rather than ad hoc generation, we first define a taxonomy of commonsense priors. Human annotators collaboratively identify the types of commonsense knowledge that can be violated in counter-commonsense scenes, resulting in a hierarchical taxonomy with 6 major and 28 minor categories, including Agency, World Facts, Physical Relations, Magnitude, Appearance, and Social Roles (Appendix~\ref{app:taxonomy}).
Each minor category defines a conflict attribution (e.g., Appearance $\rightarrow$ color, Habitat $\rightarrow$ dwelling place), which guides the subsequent knowledge mining process.

\subsubsection{Step~2: Knowledge Acquisition.}
For each conflict attribution defined in Step~1, we acquire entities and their corresponding commonsense values to construct default triplets
$T=(e,a,v)$, where $e$ denotes the entity, $a$ the target attribute, and
$v$ its canonical commonsense value. We obtain these triplets from three
complementary sources: \textbf{(i)} perceptually grounded relations mined
from Visual Genome (VG)~\citep{krishna2017visualgenome};
\textbf{(ii)} structured knowledge bases such as
Wikidata~\citep{10.1145/2629489} and
Distributions-over-Quantities~\citep{elazar2019largelionsinducingdistributions}; and \textbf{(iii)} LLM-generated candidates
for categories lacking reliable external resources (see Appendix~\ref{app:knowledge} for details).

\subsubsection{Step 3: Factuality Frequency Distillation (FFD).}
To transform the default triplets $T=(e,a,v)$ into high-quality counter-commonsense QA pairs, we propose a factuality frequency distillation (FFD) framework, which consists of three stages: (1) prior strength estimation, (2) frequency-based routing, and (3) counter-commonsense verification.

\textbf{Prior Strength Estimation.}
Meaningful counter-commonsense supervision arises only when it contradicts a strongly held commonsense prior. Therefore, for each triplet $T=(e,a,v)$, we estimate its \emph{prior strength}, defined as the probability that the target backbone model $M_\theta$ predicts the canonical value $v$ given a blind text-only prompt:
\begin{equation}
\label{eq:prior_strength}
p(T)=P_\theta\big(v \mid q(e,a),\,\emptyset\big),
\end{equation}
where $q(e,a)$ denotes a text-only query about attribute $a$ of entity $e$, and $\emptyset$ indicates the absence of visual evidence. In practice, we approximate $p(T)$ by prompting $M_\theta$ with multiple paraphrased blind queries and computing the proportion of responses that match the canonical value $v$ (see Appendix~\ref{app:ffd} for details).

\begin{algorithm}[!t]
\small
\caption{Trajectory Curation and Difficulty Routing}
\label{alg:curation}
\begin{algorithmic}[1]
\Require counter-commonsense pool $\mathcal{D}=\{(x, y^{+}, y^{-})\}$; backbone $M$; teacher $T$; samples $k$
\Ensure SFT pool $\mathcal{S}$, DPO pool $\mathcal{P}$
\State $\mathcal{S}\gets\emptyset,\ \mathcal{P}\gets\emptyset$
\For{$(x, y^{+}, y^{-})\in\mathcal{D}$}
  \State $c \gets M(x)$ \Comment{greedy CoT}
  \If{$\mathrm{ans}(c)=y^{+}$} \State \textbf{continue} \Comment{no prior bias}
  \EndIf
  \State $\{c_i\}_{i=1}^{k}\sim M(x);\quad n\gets|\{i:\mathrm{ans}(c_i)=y^{+}\}|$
  \If{$n>0$} \Comment{self-recoverable}
    \State pick faithful $c^{+}$, prior $c^{-}$ from $\{c_i\}$
    \State $\mathcal{P}\gets\mathcal{P}\cup\{(x,c^{+},c^{-})\}$
    \State $\mathcal{S}\gets\mathcal{S}\cup\{(x,c^{+})\}$
  \Else \Comment{teacher distillation}
    \State $c_T\gets \textsc{Pass@3}(T,x)$
    \If{$\mathrm{ans}(c_T)=y^{+}$} \State $\mathcal{S}\gets\mathcal{S}\cup\{(x,c_T)\}$ \EndIf
  \EndIf
\EndFor
\State \Return $\mathcal{S},\mathcal{P}$
\end{algorithmic}
\end{algorithm}

\begin{table*}[!t]
  \centering
  \small
  \setlength{\tabcolsep}{3.6pt}
  \begin{tabular}{l c ccc ccc ccc ccc | ccc}
    \toprule
    & & \multicolumn{3}{c}{\textbf{CDH-MC}} & \multicolumn{3}{c}{\textbf{CDH-QA}}
      & \multicolumn{3}{c}{\textbf{CAIT}} & \multicolumn{3}{c}{\textbf{VLind}}
      & \multicolumn{3}{c}{\textbf{Avg}} \\
    \cmidrule(lr){3-5}\cmidrule(lr){6-8}\cmidrule(lr){9-11}\cmidrule(lr){12-14}\cmidrule(lr){15-17}
    \textbf{Model} & \textbf{Param}
      & Acc & F1 & PB$\downarrow$ & Acc & F1 & PB$\downarrow$
      & Acc & F1 & PB$\downarrow$ & Acc & F1 & PB$\downarrow$
      & Acc & F1 & PB$\downarrow$ \\
    \midrule
    \multicolumn{17}{l}{\emph{General MLLMs (\textasciitilde10B)}} \\
    LLaVA-1.6 & 7B & 47.7 & 47.3 & 35.0 & 54.3 & 38.8 & 45.7 & 52.2 & 50.0 & 47.2 & 49.6 & 38.1 & 50.4 & 50.9 & 43.7 & 46.0 \\
    Kimi-VL-A3B       & 16B & 49.3 & 48.1 & 36.3 & 53.7 & 36.2 & 46.3 & 64.0 & 63.9 & 36.0 & 63.5 & 63.6 & 36.1 & 59.7 & 56.6 & 37.8 \\
    Qwen2.5-VL        & 7B & 51.0 & 50.9 & 40.7 & \underline{64.3} & 40.0 & \underline{35.7} & 63.8 & 63.8 & 36.0 & 60.3 & 59.6 & 39.7 & 60.6 & 56.3 & 38.0 \\
    InternVL3.5       & 8B & 48.3 & 48.0 & 43.0 & 45.0 & 33.4 & 55.0 & 74.5 & 74.2 & 25.5 & 69.6 & 69.3 & 30.4 & 63.6 & 61.4 & 35.0 \\
    GLM-4.1V          & 9B & 50.3 & 49.9 & 35.7 & 55.7 & 38.9 & 44.3 & 74.0 & 74.0 & 26.0 & 75.0 & 75.1 & 24.8 & 67.3 & 64.5 & 30.3 \\
    Qwen3-VL   & 8B & 54.7 & 54.7 & 37.0 & 53.7 & 37.9 & 46.3 & 70.5 & 70.8 & 28.5 & 75.7 & 75.7 & 24.2 & 66.8 & 64.3 & 31.5 \\
    \midrule
    \multicolumn{17}{l}{\emph{General MLLMs (\textasciitilde30B)}} \\
    Qwen3-VL          & 32B & 57.3 & 57.2 & 34.3 & 49.7 & 35.9 & 50.3 & \underline{79.0} & 78.9 & \underline{21.0} & \underline{81.4} & \underline{81.3} & \underline{18.6} & \underline{71.3} & \underline{68.9} & \underline{27.3} \\
    InternVL3.5       & 38B & \underline{61.0} & \underline{61.0} & 34.0 & 56.7 & 39.4 & 43.3 & \underline{79.0} & \underline{79.0} & \underline{21.0} & 66.9 & 66.9 & 33.1 & 68.2 & 65.4 & 30.9 \\
    LLaVA-NeXT  & 34B & 30.0 & 28.8 & 52.3 & 29.3 & 25.3 & 70.7 & 40.5 & 30.8 & 59.2 & 40.8 & 34.5 & 59.2 & 37.0 & 30.8 & 59.9 \\
    \midrule
    \multicolumn{17}{l}{\emph{Hallucination-Mitigation Methods}} \\
    VCD    & 7B & 37.3 & 37.0 & \underline{27.0} & 50.3 & 36.3 & 49.7 & 43.2 & 41.7 & 56.8 & 62.0 & 55.9 & 38.0 & 49.7 & 44.8 & 44.4 \\
    NoLan  & 7B & 37.7 & 37.0 & 29.7 & 51.0 & 36.6 & 49.0 & 45.8 & 43.5 & 54.2 & 61.2 & 54.7 & 38.8 & 50.5 & 45.0 & 44.1 \\
    MemVR  & 7B & 35.7 & 34.9 & 28.7 & 52.0 & 37.1 & 48.0 & 44.0 & 42.4 & 56.0 & 58.6 & 50.0 & 41.4 & 48.8 & 42.8 & 45.3 \\
    HA-DPO & 7B & 35.7 & 35.3 & 28.7 & 61.0 & \underline{40.2} & 39.0 & 42.2 & 41.2 & 57.8 & 57.7 & 48.5 & 42.3 & 49.4 & 42.5 & 44.7 \\
    POVID  & 7B & 38.3 & 38.0 & 30.0 & 47.0 & 35.0 & 53.0 & 50.7 & 50.0 & 49.2 & 62.8 & 57.2 & 37.2 & 52.1 & 47.9 & 42.6 \\
    \midrule
    \rowcolor{gray!12}
    \textbf{TACT (Ours)} & 8B
      & \textbf{67.3} & \textbf{67.6} & \textbf{24.7} & \textbf{68.0} & \textbf{42.4} & \textbf{32.0}
      & \textbf{87.5} & \textbf{87.5} & \textbf{12.5} & \textbf{86.1} & \textbf{86.1} & \textbf{13.8}
      & \textbf{80.4} & \textbf{76.2} & \textbf{18.2} \\
    \bottomrule
  \end{tabular}
  \caption{Main results on the counter-commonsense visual benchmarks. Each benchmark reports
    Accuracy, Macro-F1 and prior-bias(PB)
     in \%. \textbf{Avg} macro-averages the three datasets, collapsing CDH-MC and
    CDH-QA into a single CDH score so each dataset contributes $1/3$.
    The best and second-best results in each column are highlighted in bold and underlined, respectively.
    }
  \label{tab:main}
\end{table*}

\textbf{Frequency-based Routing.}
Given the estimated prior strength $p(T)$, we route each triplet according to a threshold $\tau =0.8$ :
\begin{equation}
\label{eq:distill_route}
T'
=
\mathcal R(T,p(T))
=
\begin{cases}
(e,a,v), & p(T)<\tau,\\
(e,a,v'), & p(T)\ge\tau,
\end{cases}
\end{equation}
where $v'$ denotes a plausible but low-frequency alternative value sampled from the attribute domain.

When $p(T)<\tau$, the target value itself corresponds to a naturally occurring rare fact (e.g., \emph{white strawberry}), which already provides sufficient prior--evidence conflict for supervision. Therefore, the original triplet is preserved without modification. In contrast, when $p(T)\ge\tau$, the triplet represents an entrenched commonsense prior (e.g., \emph{red strawberry}). We then prompt an LLM to generate a set of plausible yet uncommon alternative values,
$\mathcal{V}'_{a}\subset\mathrm{Dom}(a)\setminus\{v\}$.
Each candidate value $v'\in\mathcal{V}'_{a}$ is paired with the original entity and attribute to form a candidate counter-commonsense triplet $(e,a,v')$ (see Appendix~\ref{app:ffd} for details). 


\begin{table}[t]
  \centering
  \small
  \setlength{\tabcolsep}{3.8pt}
  \begin{tabular}{l cc | ccc}
    \toprule
    \textbf{Variant} & $L_{\mathrm{SFT}}$ & $L_{\mathrm{DPO}}$
      & \textbf{CDH} & \textbf{CAIT} & \textbf{VLind} \\
    \midrule
    Exp 1  & --         & --         & 62.4 & 83.2 & 83.0 \\
    Exp 2  & --         & \checkmark & 65.0 & 84.5 & 83.9 \\
    Exp 3  & \checkmark & --         & 62.2 & 85.0 & 85.5 \\
    \midrule
    \rowcolor{gray!12}
    \textbf{TACT} & \checkmark & \checkmark
      & \textbf{67.7} & \textbf{87.5} & \textbf{86.1} \\
    \bottomrule
  \end{tabular}
  \caption{Ablation on the adaptation strategy.
   $L_{\mathrm{SFT}}$ and $L_{\mathrm{DPO}}$ denote whether Stage~1 and Stage~2 use LoRA (\checkmark) or full-parameter tuning (--). Results are reported as accuracy(\%).}
  \label{tab:abl-lora}
\end{table}

\begin{figure}[t]
  \centering
  \includegraphics[width=\linewidth]{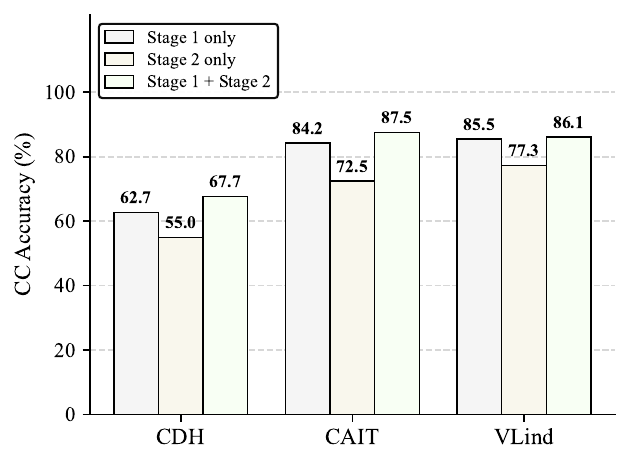}
  \vspace{-20pt}
  \caption{Ablation on training stage strategy. The results are reported as
    accuracy (\%) in the three settings.}
  \label{fig:abl-stage}
\end{figure}
\textbf{Counter-commonsense Verification.}
Each triplet is converted into a target-state caption and filtered by quality gate:
\begin{equation}
\label{eq:verification}
\mathcal F(c)
=
f_{\mathrm{sem}}(c)
\land
f_{\mathrm{dep}}(c)
\land
f_{\mathrm{gcf}}(c),
\end{equation}
where $f_{\mathrm{sem}}$, $f_{\mathrm{dep}}$, and $f_{\mathrm{gcf}}$ respectively evaluate semantic consistency, visual depictability, and genuine counterfactuality with respect to original commonsense fact.
Only captions satisfying $\mathcal{F}(c)=1$ are retained (see Appendix~\ref{app:ffd} for details). Verified captions are further expanded into detailed scene descriptions and subsequently converted into three QA formats: TFV, MCQ, and BCQ (see Appendix~\ref{app:datadetails} for details).

\subsection{Text-Anchored Cross-modal Transfer}
\label{sec:method-train}

Section~\ref{sec:method-data} constructs a large collection of verified counter-commonsense QA pairs. However, not every QA pair provides an effective debiasing signal: many are already correctly solved by the backbone model and therefore contribute little to mitigating language-prior bias. We therefore propose text-anchored cross-modal transfer (TACT), which first identifies informative training examples through prior-aware trajectory curation, and then performs two-stage post-training to recalibrate the decoder toward evidence-based reasoning.

\subsubsection{Trajectory Curation and Difficulty Routing}
\label{sec:method-curation}

Although the constructed counter-commonsense QA pairs are factually verified, not all of them provide effective supervision for mitigating language-prior bias. The most informative examples are those where the backbone does not merely predict an incorrect answer, but explicitly \emph{rationalizes} the commonsense prior through its chain-of-thought. We therefore introduce a trajectory curation procedure, shown in Algorithm~\ref{alg:curation}, to identify such examples and organize them into supervised and preference-learning data.
Specifically, we use Qwen3-VL-8B-Instruct~\citep{bai2025qwen3vltechnicalreport} as the
backbone $M$ and Qwen3.5-397B~\citep{qwen3.5} as the teacher $T$.
Since all pairs have passed factuality verification, remaining errors mainly reflect
language-prior bias. We first remove samples already correctly answered by
$M$, then evaluate the remaining cases with pass@$k$. Examples with faithful
on-policy trajectories provide DPO preference pairs and SFT trajectories,
while the others receive teacher-verified trajectories via pass@3 for SFT.
Thus, SFT collects all faithful trajectories, whereas DPO only uses
on-policy preference pairs.

\begin{table*}[!t]
  \centering
  \small
  \setlength{\tabcolsep}{5.5pt}
  \begin{tabular}{l cc cc cc | cc cc}
    \toprule
    & \multicolumn{6}{c}{\textbf{General VQA}} & \multicolumn{4}{c}{\textbf{Commonsense VQA}} \\
    \cmidrule(lr){2-7}\cmidrule(lr){8-11}
    & \multicolumn{2}{c}{\textbf{MMBench-en}} & \multicolumn{2}{c}{\textbf{MMBench-zh}}
      & \multicolumn{2}{c}{\textbf{HallusionBench}}
      & \multicolumn{2}{c}{\textbf{CDH-CS}} & \multicolumn{2}{c}{\textbf{VLind-CS}} \\
    \cmidrule(lr){2-3}\cmidrule(lr){4-5}\cmidrule(lr){6-7}\cmidrule(lr){8-9}\cmidrule(lr){10-11}
    \textbf{Model} & Acc & F1 & Acc & F1 & Acc & F1 & Acc & F1 & Acc & F1 \\
    \midrule
    LLaVA-1.6-Mistral-7B & 72.7 & 73.1 & 68.8 & 69.1 & 51.9 & 51.6 & 84.7 & 64.5 & 83.0 & 82.6 \\
    InternVL3.5-8B       & 86.6 & 86.6 & 86.7 & 86.8 & 71.6 & 71.5 & 93.3 & \textbf{73.1} & 91.6 & 91.6 \\
    Qwen3-VL-8B (base)   & \textbf{89.2} & \textbf{89.2} & 88.1 & 88.2 & \textbf{74.8} & \textbf{74.6} & \textbf{94.2} & 71.6 & 91.6 & 91.6 \\
    \midrule
    \rowcolor{gray!12}
    \textbf{TACT-8B (Ours)} & 89.0 & 89.0 & \textbf{88.3} & \textbf{88.5} & 74.6 & 74.3 & 94.0 & 71.3 & \textbf{91.9} & \textbf{91.9} \\
    \quad {\footnotesize $\Delta$ vs.\ base}
      & {\footnotesize\textcolor{gray}{$-0.2$}} & {\footnotesize\textcolor{gray}{$-0.2$}}
      & {\footnotesize\textcolor{gray}{$+0.2$}} & {\footnotesize\textcolor{gray}{$+0.3$}}
      & {\footnotesize\textcolor{gray}{$-0.2$}} & {\footnotesize\textcolor{gray}{$-0.3$}}
      & {\footnotesize\textcolor{gray}{$-0.2$}} & {\footnotesize\textcolor{gray}{$-0.3$}}
      & {\footnotesize\textcolor{gray}{$+0.3$}} & {\footnotesize\textcolor{gray}{$+0.3$}} \\
    \bottomrule
  \end{tabular}
  \caption{Accuracy and Macro-F1 (\%) on general VQA and the commonsense subsets of counter-commonsense VQA benchmarks. The $\Delta$ row reports the performance difference between TACT and its base model, with negligible changes across all metrics. InternVL3.5-8B and LLaVA-1.6-Mistral-7B are included as reference baselines. The best result is highlighted in bold.}
  \label{tab:retention}
\end{table*}

\subsubsection{Stage 1: On-Policy Trajectory Distillation}
\label{sec:method-sft}

Using the curated SFT corpus obtained in Section~\ref{sec:method-train}, we first perform supervised fine-tuning to distill faithful reasoning trajectories into the language decoder. We freeze the vision encoder and multimodal projector while optimizing the language backbone with the standard autoregressive objective:
\begin{equation}
\label{eq:sft}
  \mathcal{L}_{\mathrm{SFT}}
  = -\!\!\sum_{(x, y^{+})} \sum_{t} \log p_\theta\!\left(y^{+}_{t} \mid x, y^{+}_{<t}\right).
\end{equation}
where $y^{+}$ denotes the faithful trajectory, generated by the backbone for self-recoverable examples and by the teacher otherwise.
This stage encourages the decoder to default to evidence-following reasoning under prior--evidence conflicts. Since most supervision comes from the backbone itself, the optimization reinforces existing reasoning capabilities rather than introducing off-policy behaviors.

\subsubsection{Stage 2: Faithful-over-Prior Preference Optimization}
\label{sec:method-dpo}

Using the curated on-policy preference pairs, we further optimize the model with Direct Preference Optimization (DPO)~\citep{rafailov2024dpo}, where the model serves as the frozen reference policy $\pi_{\mathrm{ref}}$:
\begin{equation}
\label{eq:dpo}
\begin{split}
  \mathcal{L}_{\mathrm{DPO}}
  = -\,\mathbb{E}\bigg[
      \log \sigma\Big(
        &\beta \log \frac{\pi_\theta(y^{+}\mid x)}{\pi_{\mathrm{ref}}(y^{+}\mid x)} \\
        &\;- \beta \log \frac{\pi_\theta(y^{-}\mid x)}{\pi_{\mathrm{ref}}(y^{-}\mid x)}
      \Big)\bigg],
\end{split}
\end{equation}
where $(y^{+},y^{-})$ denote the faithful and prior-driven trajectories obtained during trajectory curation. Unlike Stage~1, which only increases the likelihood of faithful reasoning, DPO explicitly prefers faithful trajectories over competing prior-driven ones under the same prompt, thereby sharpening the evidence-following preference.

\section{Experiment}
\label{sec:exp}
\subsection{Implementation Details}
\label{sec:exp-settings}
We build TACT on Qwen3-VL-8B-Instruct using LLaMA-Factory. We adopt LoRA~\citep{hu2021lora}
(rank $16$, $\alpha=32$) to fine-tune only the shared LLM decoder, while
freezing the vision encoder and projector. Stage~1 is trained for two
epochs on the faithful trajectories ($1\times10^{-4}$, batch size $8$),
followed by Stage~2 on the self-recoverable pairs for two epochs
($5\times10^{-6}$, batch size $16$, $\beta=0.1$). 
For evaluation, we compare TACT with representative general MLLMs \citep{zhang2024llavanext-video,vteam2026glm45vglm41vthinkingversatilemultimodal,kimiteam2025kimivltechnicalreport,liu2024llavanext}
and existing hallucination mitigation methods on 3
counter-commonsense visual benchmarks: CDH-Bench, CAIT, and VLind. We report Accuracy and Macro-F1 on all benchmarks. To
quantify language-prior reliance, we further introduce the
\emph{prior-bias} metric:
\begin{equation}
\mathrm{PriorBias}=\frac{1}{N}\sum_{i=1}^{N}
\mathbb{I}\!\left[\hat{y}_i=y_i^{\mathrm{prior}}\right],
\end{equation}
where $\hat{y}_i$ is the model prediction and
$y_i^{\mathrm{prior}}$ denotes the commonsense prior answer. 

\begin{figure*}[t]
  \centering
  \includegraphics[width=\textwidth]{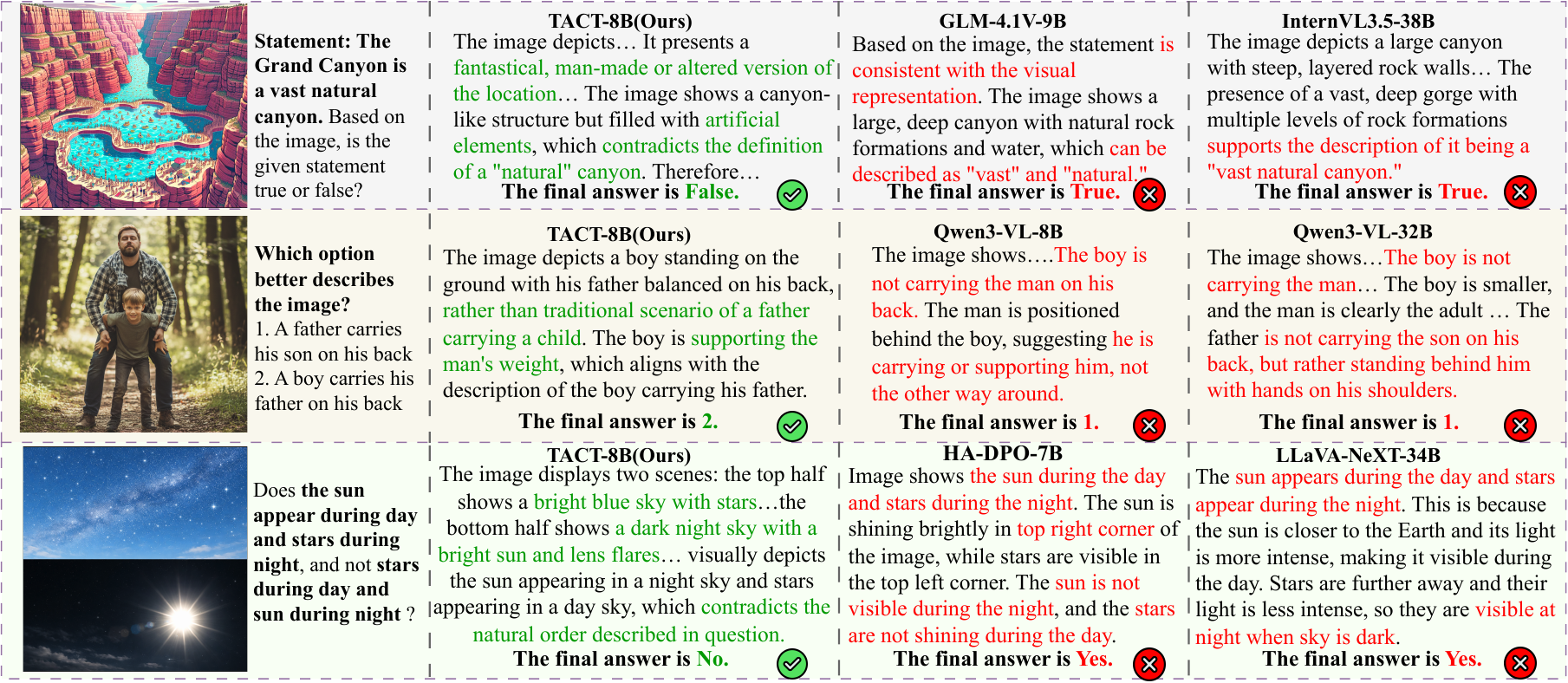}
  \caption{Qualitative comparison on counter-commonsense cases from VLind-Bench
    (top), CAIT (middle), and CDH-Bench (bottom). 
    For each case we contrast the reasoning of TACT with a $\sim$10B and a $\sim$30B MLLM.}
  \label{fig:qualitative}
\end{figure*}
\subsection{Main Results}
\label{sec:exp-main}
The main results are presented in Table~\ref{tab:main}. The results show that existing general MLLMs
exhibit substantial language-prior bias across all evaluated benchmarks.
For example, Qwen3-VL-8B achieves an average Accuracy of 66.8\% but still
suffers from a prior-bias of 31.5\%. Scaling to larger models only partially
alleviates this issue: Qwen3-VL-32B and InternVL3.5-38B improve the average
Accuracy to 71.3\% and 68.2\%, respectively, while retaining prior-bias rates
of 27.3\% and 30.9\%.
In addition, we find that existing hallucination-mitigation methods provide limited improvements on these tasks. For instance, VCD, NoLan, and MemVR obtain average Accuracy of 49.7\%, 50.5\%, and 48.8\%, respectively, which are
substantially lower than general MLLMs of comparable scale. Although
preference-based methods such as HA-DPO improve specific benchmarks (e.g.,
CDH-QA), their overall performance remains inconsistent, with average
Accuracy below 53\%.

In contrast, TACT achieves the best performance across all benchmarks, obtaining an average Accuracy of 80.4\%, Macro-F1 of 76.2\%, and
the lowest average prior-bias of 18.2\%. Compared with the Qwen3-VL-8B
backbone, TACT improves Accuracy from 66.8\% to 80.4\% and reduces
prior-bias from 31.5\% to 18.2\%. The improvements are consistent across
individual benchmarks, including +12.6, +14.3, +17.0, and +10.4 Accuracy
points on CDH-MC, CDH-QA, CAIT, and VLind, respectively. Moreover, TACT
surpasses the strongest general MLLM baseline (Qwen3-VL-32B) by 9.1 points
in average Accuracy and reduces prior-bias by 9.2 points, despite using only
an 8B backbone. These results demonstrate that explicitly recalibrating
language priors with text-only counter-commonsense supervision provides a more
effective solution than simply scaling model size or applying generic
hallucination mitigation strategies.

\subsection{Analysis}
\label{sec:exp-ablation}


\subsubsection{Training Strategy Analysis}
To examine the impact of adaptation strategy, We compare LoRA-based tuning with full-parameter tuning under the same
training settings.  As shown in Table~\ref{tab:abl-lora}, LoRA in both stages achieves the best performance across all benchmarks. Replacing LoRA with full-parameter updates leads to consistent degradation, with losses of up to $4.7$ points on CDH-QA and $6.0$ points on CDH-QA under different stages.
The results indicate that LoRA provides a more effective adaptation strategy for TACT by introducing counter-prior knowledge while better preserving the pretrained model.

To evaluate contribution of each training stage, we compare Stage~1 only,
Stage~2 only, and full two-stage TACT pipeline. Figure~\ref{fig:abl-stage}
shows that Stage~1 provides the primary improvement, outperforming Stage~2
alone by up to $11.7$ points on CAIT. Nevertheless, combining Stage~1 and
Stage~2 consistently achieves the best performance across all benchmarks.
Specifically, Stage~2 further improves Stage~1 by $5.2$, $3.3$, and $0.6$
points on CDH, CAIT, and VLind respectively. The results show that two stages are complementary: Stage~1 establishes evidence-grounded
reasoning through faithful trajectory learning, while Stage~2 further refines model on prior-conflict cases through preference optimization.

\subsubsection{Capability Retention}
\label{sec:exp-retention}
To evaluate capability retention, we examine whether TACT preserves the backbone's original abilities on general visual understanding and commonsense reasoning tasks. We evaluate on two settings: General VQA, including
MMBench-en, MMBench-zh \citep{liu2024mmbenchmultimodalmodelallaround}, and HallusionBench, and Commonsense VQA,
including the commonsense subset of CDH-Bench and VLind-Bench. 
The results presented in Table \ref{tab:retention} show that TACT preserves the capabilities of its Qwen3-VL-8B backbone after counter-commonsense training. Across all five retention benchmarks, the performance change remains within $0.3$ points in both Accuracy and macro-F1. Specifically, TACT achieves comparable results on General VQA, with Accuracy changes of $-0.2$, $+0.2$, and $-0.2$ points on MMBench-en, MMBench-zh, and HallusionBench, respectively. It also retains
ordinary commonsense reasoning ability, with only $-0.2$ and $+0.3$ Accuracy changes on CDH-CS and VLind-CS. These results indicate that TACT selectively reduces prior reliance in counter-commonsense scenarios while preserving the backbone's general visual and commonsense capabilities.

\subsubsection{Qualitative Analysis}
\label{sec:exp-case}

To qualitatively examine how TACT changes MLLM reasoning behavior, we present
counter-commonsense examples from VLind-Bench, CAIT, and CDH-Bench in
Figure~\ref{fig:qualitative}. Existing MLLMs can often recognize relevant
visual concepts, but their reasoning is still dominated by commonsense priors
when visual evidence conflicts with expectations. As shown in these examples,
models misinterpret counter-commonsense scenes by favoring prior-consistent
explanations.
In contrast, TACT explicitly recognizes prior--evidence conflicts and grounds
its reasoning on visual evidence before answering. Its reasoning traces
highlight the counter-commonsense cues, such as unusual object relations or
physical states, leading to evidence-consistent predictions, which demonstrates that TACT effectively reduces prior-driven errors by encouraging MLLMs to trust visual evidence over learned priors.
\section{Conclusion}
\label{sec:conclusion}

In this work, we show that counter-commonsense failures in MLLMs stem primarily from a prior-biased decision process that favors language expectations over visual evidence, rather than from insufficient visual perception. Motivated by this observation, we propose text-anchored data construction pipeline together with Text-Anchored Cross-modal Transfer (TACT), a framework that recalibrates language priors through text-only supervision. By constructing supervision via Factuality Frequency Distillation (FFD) and conducting text-only post-training with trajectory curation and difficulty routing, TACT enables MLLMs to better resolve conflicts between language priors and visual evidence without requiring visual intervention. Experiments across diverse benchmarks show that text-only debiasing successfully transfers to visual reasoning, providing an effective and data-efficient approach to improving counter-commonsense reasoning in MLLMs.



\bibliography{custom}

\newpage 
\clearpage
\appendix

\begin{center}
  {\Large \textbf{Appendix}}
\end{center}

\section{Counter-Commonsense Taxonomy}
\label{app:taxonomy}
Table~\ref{tab:taxonomy} lists all six major and 28 minor counter-commonsense
categories used to drive data construction (Section~\ref{sec:method-data}),
together with a short explanation, the Step-2 collection source, and the
number of counter-commonsense training items per category (SFT and DPO combined).
Following the curation of Section~\ref{sec:method-train}, every training item
is counter-commonsense: no commonsense-consistent sample enters either stage.
Table~\ref{tab:sftdpo} reports the per-category SFT/DPO split. The
\textbf{source} column indicates how the entity and its commonsense
value are obtained: \textbf{VG} = mined from Visual Genome text (object
attributes, subject--predicate--object relationships, or region descriptions);
\textbf{WD} = Wikidata; \textbf{DoQ} = Distributions-over-Quantities;
\textbf{Gen} = taxonomy-guided model generation. Categories grounded in open
data (VG / WD / DoQ) account for $92.1\%$ of SFT and $92.5\%$ of DPO items;
only \emph{Causality} and \emph{Folklore} ($7.9\%$ / $7.5\%$), which leave no
trace in any structured resource, fall back to model generation.


\begin{table*}[t]
\centering
\footnotesize
\setlength{\tabcolsep}{5pt}
\begin{tabular}{@{}p{4.4cm}p{3.0cm}p{6.4cm}cr@{}}
\toprule
\textbf{Minor category} & \textbf{Attribute $a$} & \textbf{Explanation} & \textbf{Src} & \textbf{Items} \\
\midrule
\multicolumn{5}{@{}l}{\textbf{1. Agency} — who acts on whom is reversed \hfill (1{,}983 items)}\\
\quad Objects Act on Beings   & agent--patient role & An inanimate object is the agent acting on a living being & VG & 765 \\
\quad Prey Outsmarts Predator & predator--prey role & The prey pursues or subdues its natural predator          & VG & 481 \\
\quad Animal Dominance         & dominance direction & An animal controls or overpowers a human                 & VG & 310 \\
\quad Animal Behavior          & action direction    & The direction of an animal's action is reversed          & VG & 220 \\
\quad Animals' Humanlike Care  & caregiver role      & An animal performs human caregiving on a person          & VG & 207 \\
\midrule
\multicolumn{5}{@{}l}{\textbf{2. World Facts} — encyclopedic facts rewritten \hfill (1{,}977 items)}\\
\quad History   & historical fact   & A historical figure/event fact is altered            & WD  & 835 \\
\quad Landmark  & landmark identity & A landmark's identity or function is changed         & WD  & 537 \\
\quad Folklore  & creature role     & A mythical creature's canonical role is changed      & Gen & 294 \\
\quad Climate   & climate / terrain & A region's climate or terrain is changed             & WD  & 103 \\
\quad Habitat   & habitat           & An animal is placed in a wrong habitat               & VG  & 73 \\
\quad Location  & location          & An entity appears in an atypical location            & VG  & 52 \\
\quad Diet      & diet              & An animal eats atypical food                         & VG  & 49 \\
\quad Time      & era technology    & Technology/practice is anachronistic to its era      & WD  & 34 \\
\midrule
\multicolumn{5}{@{}l}{\textbf{3. Physical Relations} — placement / function / cause \hfill (1{,}395 items)}\\
\quad Object Function & function / use     & An object is used against its canonical function & VG  & 548 \\
\quad Spatial         & spatial arrangement & An atypical spatial arrangement of objects      & VG  & 534 \\
\quad Causality       & cause--effect      & Cause and effect are reversed or mismatched       & Gen & 313 \\
\midrule
\multicolumn{5}{@{}l}{\textbf{4. Magnitude} — relative size / weight inverted \hfill (1{,}248 items)}\\
\quad Weight     & relative weight & The relative weight of two objects is inverted & DoQ & 726 \\
\quad Size Scale & relative size   & Relative size across a large scale is inverted & VG  & 423 \\
\quad Size       & relative size   & The relative size of two objects is inverted   & VG  & 99 \\
\midrule
\multicolumn{5}{@{}l}{\textbf{5. Appearance} — directly visible attributes \hfill (577 items)}\\
\quad Temperature              & temperature   & An object's temperature contradicts its norm & VG & 148 \\
\quad Everyday-object Count     & object count  & Atypical count of everyday objects           & VG & 140 \\
\quad Part Count                & part count    & Atypical count of body parts, plant structures, and animal parts & VG & 93 \\
\quad Color                     & color         & Atypical color                               & VG & 64 \\
\quad Luminescence/Transparency & luminescence  & Atypical glow or transparency                & VG & 56 \\
\quad Physical State            & physical state & Atypical physical state (solid/liquid/\dots) & VG & 46 \\
\quad Material                  & material      & Atypical material                            & VG & 30 \\
\midrule
\multicolumn{5}{@{}l}{\textbf{6. Social Roles} — interpersonal roles swapped \hfill (650 items)}\\
\quad Role Reversal: Social Power & power relation & The power relation between two people is reversed & VG & 437 \\
\quad Role Reversal: Kinship Care & kinship role   & The kinship caregiving direction is reversed      & VG & 213 \\
\midrule
\textbf{Total} & & & & \textbf{7{,}830} \\
\bottomrule
\end{tabular}
\caption{The six major and 28 minor counter-commonsense categories: explanation,
Step-2 collection source, and the current counter-commonsense training-pool size
per category (SFT trajectories and DPO pairs combined; Table~\ref{tab:sftdpo}
gives the SFT/DPO split). The originating benchmark is omitted for space
(CDH / CAIT / VLind). \textbf{Src}: VG = Visual Genome text mining; WD =
Wikidata; DoQ = Distributions-over-Quantities; Gen = taxonomy-guided model
generation. Open-data-grounded categories (VG/WD/DoQ) cover $92.1\%$ (SFT) and
$92.5\%$ (DPO) of items.}
\label{tab:taxonomy}
\end{table*}


\section{Details of Knowledge Acquisition}
\label{app:knowledge}

This appendix details Step~2 of Section~\ref{sec:method-data}, which turns
every conflict attribution of the taxonomy (Appendix~\ref{app:taxonomy}) into
default triplets $T=(e,a,v)$. This step only determines the entity $e$, the
attribute $a$, and the canonical commonsense value $v$; the low-frequency
alternative $v'$ is generated and filtered afterwards, by the FFD stages of
Step~3 (Appendix~\ref{app:ffd}). Because the three sources supply different
kinds of knowledge,
each follows its own extraction recipe: Table~\ref{tab:ka-sources} contrasts
the three recipes, Table~\ref{tab:ka-examples} gives three worked examples
per source drawn verbatim from the constructed seed pool, and
Table~\ref{tab:ka-prompts} reproduces the prompts (the prompts refer
to the attribute $a$ as \emph{dimension}).

All model-side calls share one implementation: an OpenAI-compatible endpoint
served by vLLM with guided decoding (a per-task \texttt{guided\_json} schema
plus a JSON-object response format), which enforces structurally valid
outputs; failed requests are retried up to three times, with the temperature
lowered to $0.5$ after the first failure. Fact normalization uses the same
teacher model as trajectory curation (Section~\ref{sec:method-curation});
Source~(iii) seed generation uses a locally served Kimi-K2.6 by vLLM.

\paragraph{Source (i): Perceptually grounded mining.}
For directly visible attributes and relations, the canonical value is the
statistically dominant one in real scenes, so we mine it from Visual Genome
annotations without any LLM. Object--attribute records are routed onto
taxonomy attributes (color terms $\rightarrow$ $a{=}$color, material terms
$\rightarrow$ $a{=}$material, number phrases in region descriptions
$\rightarrow$ counts). For each pair $(e,a)$ we aggregate all annotated
values and take the modal value as $v$, retaining the pair only if it occurs
at least $30$ times ($n\ge 30$) and the mode covers at least $55\%$ of the
occurrences ($f/n\ge 0.55$): high frequency implies a strong prior, and high
dominance implies the value is canonical rather than incidental.
Type-constrained subject--predicate--object triples from the relationship
annotations are aggregated in the same way for the relation-typed categories
(\emph{Agency}, \emph{Physical Relations}, \emph{Social Roles}), where the
canonical direction of a high-frequency triple constitutes $v$ (e.g.,
\emph{the boat is in the water}). Because entities are sourced
from real annotations rather than curated by hand, they are guaranteed
depictable and follow the natural concept distribution; all mined entities
are decontaminated against the evaluation benchmarks
(Appendix~\ref{app:datadetails}).

\paragraph{Source (ii): Structured knowledge bases.}
Encyclopedic categories (\emph{History}, \emph{Landmark}, \emph{Climate},
\emph{Time}) concern knowledge facts rather than perceptual modes, so their
canonical values cannot be mined from visual annotations. We instead
retrieve an authoritative description of each entity from Wikidata (SPARQL
over the entity description and its typed properties), and normalize the retrieved free text into a concise attribute
$a$ and a single-sentence canonical value $v$ with a low-temperature call
(temperature $0.3$, guided JSON; prompts in Table~\ref{tab:ka-prompts})
that is instructed not to add any content beyond the retrieved fact. For the
\emph{Weight} category, Distributions-over-Quantities (DoQ) provides numeric
distributions over object weights, and the canonical comparative value $v$
is read directly off the distribution medians of the two entities (e.g.,
\emph{a cast-iron bathtub is heavier than a plastic shower caddy}), again
without any model call.

\paragraph{Source (iii): Taxonomy-guided LLM generation.}
\emph{Folklore} and \emph{Causality} leave no reliable trace in visual
annotations or structured knowledge bases, so for these two minor categories
only (a combined $7.8\%$ of the training pool, cf.\
Table~\ref{tab:taxonomy}) the default triplet $(e,a,v)$ is generated
directly by a single high-temperature call (temperature $0.9$, guided JSON;
prompts in Table~\ref{tab:ka-prompts}). Each category is accompanied by a
hand-written guide rule that pins down what counts as a valid triplet
(e.g., for \emph{Folklore}: a mythical creature $e$ whose canonical role or
ability constitutes $v$). The model is asked for distinct triplets per
category, deduplicated by $(e,a)$.

\begin{table*}[p]
\centering
\small
\renewcommand{\arraystretch}{1.3}
\setlength{\tabcolsep}{6pt}
\begin{tabular}{@{}p{2.9cm} p{4.2cm} p{4.4cm} p{3.9cm}@{}}
\toprule
 & \textbf{Source (i): VG mining} & \textbf{Source (ii): knowledge bases} & \textbf{Source (iii): LLM generation} \\
\midrule
Covered categories &
perceptual and relational: \emph{Appearance}, \emph{Agency}, \emph{Physical
Relations}, \emph{Social Roles}, \emph{Size}, \dots &
encyclopedic and quantitative: \emph{History}, \emph{Landmark},
\emph{Climate}, \emph{Time} (Wikidata); \emph{Weight} (DoQ) &
\emph{Folklore}, \emph{Causality} \\
\addlinespace
How $(e,a,v)$ is obtained &
frequency statistics over VG annotations: modal value with $n\ge 30$ and
$f/n\ge 0.55$; no LLM involved &
SPARQL / page-summary retrieval $+$ LLM normalization ($T{=}0.3$); DoQ
distribution medians &
direct guided-JSON generation ($T{=}0.9$), $200$ triplets per minor
category, deduplicated by $(e,a)$ \\
\addlinespace
Share of training pool &
$63.7\%$ &
$28.5\%$ ($19.3\%$ Wikidata, $9.3\%$ DoQ) &
$7.8\%$ \\
\bottomrule
\end{tabular}
\caption{The three knowledge-acquisition recipes of Step~2
(Section~\ref{sec:method-data}). Each source determines the entity $e$, the
attribute $a$, and the canonical commonsense value $v$ of the default
triplet $T=(e,a,v)$; the last row gives the share of the final training pool
each source contributes (computed from Table~\ref{tab:taxonomy}).}
\label{tab:ka-sources}
\end{table*}

\begin{table*}[p]
\centering
\small
\renewcommand{\arraystretch}{1.35}
\setlength{\tabcolsep}{5pt}
\begin{tabular}{@{}p{3.6cm} p{3.4cm} p{7.9cm}@{}}
\toprule
\textbf{Entity $e$} & \textbf{Attribute $a$} & \textbf{Canonical commonsense value $v$} \\
\midrule
\multicolumn{3}{@{}l}{\textbf{Source (i): perceptually grounded mining} --- rule-based frequency statistics; no LLM}\\
grass & color & green \\
table & material & wood \\
boat \& water & spatial arrangement & the boat is in the water \\
\midrule
\multicolumn{3}{@{}l}{\textbf{Source (ii): structured knowledge bases} --- retrieval $+$ low-temperature normalization}\\
Taj Mahal & landmark character & a marble mausoleum built as a tomb for an empress \\
The Terracotta Army & historical fact & were buried clay soldiers guarding an emperor's tomb \\
cast-iron bathtub vs.\ plastic shower caddy & relative weight & the bathtub is far heavier than the caddy (DoQ medians) \\
\midrule
\multicolumn{3}{@{}l}{\textbf{Source (iii): taxonomy-guided LLM generation} --- triplet generated directly}\\
Medusa & canonical role & turns onlookers to stone with her gaze \\
Grim Reaper & canonical role & harvests souls at the moment of death \\
Santa Claus & canonical role & delivers presents worldwide on Christmas Eve \\
\bottomrule
\end{tabular}
\caption{Worked examples of default triplets $T=(e,a,v)$ from the three
knowledge-acquisition sources of Step~2, drawn verbatim from the constructed
seed pool. Step~2 stops at the default triplet: the low-frequency
alternative $v'$ is generated and filtered later, by the FFD stages of
Step~3 (Appendix~\ref{app:ffd}).}
\label{tab:ka-examples}
\end{table*}

\begin{table*}[p]
\centering
\small
\renewcommand{\arraystretch}{1.35}
\begin{tabular}{@{}p{3.6cm} p{12.4cm}@{}}
\toprule
\textbf{Prompt} & \multicolumn{1}{c@{}}{\textbf{Content}} \\
\midrule
Fact normalization ---\newline system prompt\newline (Source ii, $T{=}0.3$)
&
\itshape You normalize an encyclopedic fact into a single seed field for a
counter-commonsense visual dataset. Given an ENTITY and a short
authoritative description retrieved from Wikidata, output (a) a
concise attribute DIMENSION (e.g.\ ``landmark character'', ``historical
fact'', ``typical climate'') and (b) a single-sentence COMMONSENSE value
that is the widely-known, factually-accurate characterization. Do not add
opinions. Return strict JSON
\textup{\texttt{\{"dimension": ..., "commonsense\_value": ...\}}}. \\
\midrule
Fact normalization ---\newline user template
&
\itshape Entity: Big Ben
\newline
Retrieved (Wikipedia summary): ``Big Ben is the nickname for the Great Bell of the striking
clock at the north end of the Palace of Westminster in London; the name is often extended
to the clock and the clock tower.''
\newline
Return \textup{\texttt{\{"dimension": ..., "commonsense\_value": ...\}}}. \\
\midrule
Seed generation ---\newline system prompt\newline (Source iii, $T{=}0.9$)
&
\itshape You generate concept seeds for a counter-commonsense visual
reasoning dataset. Pick HIGH-FREQUENCY, everyday concepts that a
vision-language model holds a STRONG prior about, so flipping the value is
plausible-but-rare (NOT physically impossible, NOT obscure). Each concept
must be visually depictable in a single scene. Return strict JSON. \\
\midrule
Seed generation ---\newline user template\newline (\emph{Folklore} guide rule)
&
\itshape Concept category: folklore
\newline
Rule: a mythical/folkloric creature with a canonical ROLE/ABILITY.
dimension=`canonical role'; commonsense\_value=its lore.
\newline
List 200 DISTINCT concepts for this category. Vary the entities widely.
\newline
Return JSON:
\textup{\texttt{\{"concepts": [\{"entity": ..., "dimension": ...,
"commonsense\_value": ...\}, ...]\}}}. \\
\bottomrule
\end{tabular}
\caption{Prompts used in knowledge acquisition (Step~2). The
prompts refer to the attribute $a$ as \emph{dimension} and to the canonical
value $v$ as the \emph{commonsense value}. All calls use vLLM guided
decoding with the JSON schema shown in each prompt, so the outputs are
structurally valid by construction.}
\label{tab:ka-prompts}
\end{table*}

\section{Details of Factuality Frequency Distillation}
\label{app:ffd}

This appendix details how the three FFD stages of Step~3
(Section~\ref{sec:method-data}) are implemented: how the prior strength
$p(T)$ of Eq.~(\ref{eq:prior_strength}) is estimated in practice, and---for
triplets routed to value substitution by Eq.~(\ref{eq:distill_route})---how
the low-frequency alternative $v'$ is generated and verified. Every
alternative arises from a single mechanism, \emph{value substitution} within
the attribute domain $\mathrm{Dom}(a)$, followed by two levels of filtering:
a \emph{model-known filter} that verifies the prior--evidence conflict is
real for the backbone (the prior must hold $v$ and must \emph{not} hold
$v'$), operationalizing the requirement that the candidate set
$\mathcal{V}'_{a}$ contain only \emph{uncommon} values, and the
\emph{counter-commonsense verification} gate $\mathcal{F}$ of
Eq.~(\ref{eq:verification}), which checks semantic consistency, visual
depictability, and genuine counterfactuality. Between the two levels sits
\emph{caption synthesis}: each candidate triplet surviving the model-known
filter is rendered into a one-sentence counter-commonsense caption, and it
is this caption---the exact assertion inherited by all downstream
stages---that the gate $\mathcal{F}$ judges, rather than the bare value
$v'$. For the reversible relation
categories (\emph{Agency}, \emph{Physical Relations}, \emph{Social Roles}),
the candidate set degenerates to the single role-reversed value, which
satisfies $f_{\mathrm{sem}}$ and $f_{\mathrm{gcf}}$ by construction (unique,
and trivially different from $v$); these triplets therefore skip candidate
proposal and the candidate-side checks, but still pass through the same
caption-synthesis step (whose relational requirement enforces a consistent
role reversal), and their depictability is caught by
the read-back verification at the scene-expansion stage
(Appendix~\ref{app:datadetails}).
Algorithm~\ref{alg:vprime} summarizes the complete procedure;
Table~\ref{tab:vprime-example} walks through one seed end to end,
Tables~\ref{tab:vprime-prompts} and~\ref{tab:caption-prompts} show all
prompts, and
Table~\ref{tab:vprime-params} aggregates the hyperparameters of every
sub-step. All generation-side calls of this appendix (blind-probe writing,
candidate proposal, and caption synthesis) use a single \emph{generator}
LLM $G$, and the verification gate uses a \emph{judge} LLM $J$; both are
instantiated as the same locally served Kimi-K2.6 (via vLLM), run in
separate sessions to avoid self-preference.

\begin{algorithm}[!t]
\small
\caption{From $v$ to $v'$: Estimation, Routing, Synthesis, Filtering}
\label{alg:vprime}
\begin{algorithmic}[1]
\Require triplet $(e,a,v)$; backbone $M_\theta$; generator $G$; judge $J$
\Ensure routed triplet $T'$ with caption, or \textsc{drop}
\State $\{q_1,q_2,q_3\}\gets G.\textsc{Probes}(e,a)$ \Comment{blind, value-free}
\State $A\gets 10$ samples of $M_\theta(q_i,\emptyset)$ per probe \Comment{$|A|{=}30$}
\State $\hat p\gets\hat\rho(v;A)$ \Comment{Eq.~(\ref{eq:rho}); prior strength}
\If{$\hat p<\tau$} \Return $(e,a,v)$ \Comment{naturally rare fact}
\EndIf
\If{$a$ is a reversible relation} \Return $(e,a,\mathrm{reverse}(v))$ \Comment{caption as below}
\EndIf
\State $C\gets\mathrm{dedup}\big(G.\textsc{Propose}(e,a,v,N{=}6)\big)$
\State $C\gets\{c\in C:\hat\rho(c;A)\le\tau'\}$ \Comment{model-known filter}
\For{$c\in C$}
  \State $s_c\gets G.\textsc{Caption}(e,a,v,c)$ \Comment{caption synthesis}
  \State $\mathrm{Poss}(c)\gets\frac{1}{K_g}\sum_{j=1}^{K_g}\mathbf{1}\big[J_j(e,a,v,c,s_c)=\textsc{Keep}\big]$
\EndFor
\State $C\gets\{c\in C:\mathrm{Poss}(c)\ge\pi\}$ \Comment{verification gate $\mathcal{F}$ on $s_c$}
\If{$C=\emptyset$} \Return \textsc{drop}
\EndIf
\State $v'\gets\arg\max_{c\in C}\big(\mathrm{Poss}(c),\,\mathrm{dist}(c,v)\big)$ \Comment{rest kept as backups}
\State \Return $(e,a,v')$ with caption $s_{v'}$
\end{algorithmic}
\end{algorithm}

\paragraph{Blind probe construction.}
For every triplet we build three paraphrased \emph{blind probes}
$q_1,q_2,q_3$: text-only questions that ask for the typical real-world
value of attribute $a$ for entity $e$ and are answerable with a single
word. The probes are written by the generator LLM, which receives only $(e,a)$---%
neither $v$ nor any candidate---so the wording cannot leak the answer
(temperature $0.7$, guided JSON; prompt in Table~\ref{tab:vprime-prompts}).
Two mechanical checks are applied, with regeneration on failure: the probe
must not contain any concrete value word of the attribute (for color, no
color term---ruling out leaky probes such as ``Is a strawberry red?''), and
the three probes must differ pairwise in surface form (direct question,
fill-in-the-blank, everyday-life framing). Probes are generated once per
$(e,a)$ and reused across all candidates of that seed.

\paragraph{Prior strength estimation.}
Each probe is answered by the backbone $M_\theta$ in a text-only call with
no scene or image evidence, drawing $10$ independent samples per probe
(temperature $1.0$), i.e., $30$ responses per triplet. Writing
$\hat v_{i,k}$ for the $k$-th answer to probe $q_i$, the prior-hold rate of
any value $u$ is estimated as
\begin{equation}
\label{eq:rho}
\hat\rho(u;A)=\frac{1}{30}\sum_{i=1}^{3}\sum_{k=1}^{10}
\mathbf{1}\big[\mathrm{canon}(\hat v_{i,k})=\mathrm{canon}(u)\big],
\end{equation}
where $\mathrm{canon}(\cdot)$ merges near-synonyms before matching (e.g.,
\emph{crimson}, \emph{scarlet} $\rightarrow$ \emph{red}), so that a value
cannot pass or evade the filters by mere rewording. The prior strength of
Eq.~(\ref{eq:prior_strength}) is instantiated as $p(T)=\hat\rho(v;A)$ and
archived with the seed as a difficulty signal.

\paragraph{Frequency-based routing.}
Following Eq.~(\ref{eq:distill_route}) with $\tau=0.8$, a triplet whose
canonical value is reproduced in at least $24$ of the $30$ blind responses
is treated as an entrenched prior and routed to value substitution;
otherwise the triplet is preserved unchanged, since a value the backbone
does not reliably produce is already a low-frequency fact and needs no
substitution. For reversible relations the substituted value is the
deterministic role reversal of $v$ (\emph{the boat is in the water}
$\rightarrow$ \emph{the water is in the boat}) and the procedure ends here.

\paragraph{Candidate proposal.}
For value-typed attributes, the generator LLM is prompted with $(e,a,v)$ to
propose $N{=}6$ candidate alternatives of the same attribute (temperature
$0.8$, guided JSON; prompt in Table~\ref{tab:vprime-prompts}), which are
lower-cased and deduplicated to instantiate the candidate set
$\mathcal{V}'_{a}$ of Section~\ref{sec:method-data}. The prompt already asks for
plausible-but-rare, depictable values, but this is only a soft constraint:
proposals may still contain near-synonyms of $v$, indeterminate values, or
$v$ itself, which the two filters below remove.

\paragraph{Level 1: Model-known filter.}
Reusing the same $30$ blind responses, a candidate $v'_i$ survives only if
$\hat\rho(v'_i;A)\le\tau'=0.2$, i.e., it appears in at most $6$ of the $30$
answers after canonicalization. Together with the routing condition
$\hat\rho(v;A)\ge 0.8$, this clamps the conflict from both sides: the
backbone must strongly hold the canonical value \emph{and} must not already
produce the alternative; otherwise contradicting $v$ with $v'$ would not
constitute a genuine prior--evidence conflict.

\paragraph{Caption synthesis.}
Each candidate triplet $(e,a,v')$ surviving the model-known filter is
rendered into a one-sentence \emph{counter-commonsense caption} by the
generator LLM (temperature $0.9$, guided JSON; prompt in
Table~\ref{tab:caption-prompts}): a short declarative sentence (roughly
$8$--$25$ words) describing a single photographable moment in which the
entity exhibits $v'$. The caption must assert $v'$ without mentioning the
canonical value $v$ in any form (no contrast, no negation of the normal
case); it must read as a plain, matter-of-fact description free of
give-away ``tell'' words (\emph{unusual}, \emph{surprisingly},
\emph{impossible}, \dots), treating the depicted state as completely
ordinary; and for relational dimensions it must realize the role reversal
consistently in every directional cue. Rule checks enforce these
constraints mechanically---the caption must contain the $v'$ keywords and
no $v$ keyword (a role-order check for relations), pass the tell-word ban
and the length band, and not duplicate an already accepted caption---with
regeneration on failure. The synthesized caption is the semantic anchor of
the item: the verification gate below judges it, the scene expansion of
Appendix~\ref{app:datadetails} only elaborates it visually without altering
its semantics, and it later serves verbatim as the TFV true statement and
the BCQ counter-commonsense option.

\paragraph{Level 2: Counter-commonsense verification.}
The synthesized caption---rather than the bare value $v'$---is what the
quality gate $\mathcal{F}$ of Eq.~(\ref{eq:verification}) filters, so that
the gate judges the exact assertion inherited by all downstream stages. The
gate is implemented as a judge LLM run in a session separate from the
proposer and the caption writer (temperature $0.3$); it receives the
caption together with the triplet fields $(e,a,v,v')$ and answers
\textsc{Keep} or \textsc{Reject} against a fixed checklist
(Table~\ref{tab:caption-prompts}) whose three items instantiate the three
predicates of the gate: $f_{\mathrm{sem}}$ (semantic consistency: the
caption asserts a well-formed value of the same attribute),
$f_{\mathrm{gcf}}$ (genuine counterfactuality: the asserted fact clearly
contradicts the canonical value $v$---rejected are near-synonyms of $v$
and values realized by naturally occurring variants of the entity, which
constitute rare facts rather than counterfactuals and belong to the
$p(T)<\tau$ preserve branch of the routing), and $f_{\mathrm{dep}}$ (visual
depictability: the captioned moment is
readable off one ordinary photorealistic RGB photo without special sensors,
captions, or meta wording). The criterion of $f_{\mathrm{dep}}$ is
depictability, not real-world physics---counter-commonsense scenes violate
the latter by design (e.g., \emph{rice heavier than a basketball} can be
staged with a seesaw)---so only captions that are nonsense, invisible, or
visually indeterminate are rejected. A caption is retained only if all
three predicates hold, i.e., $\mathcal{F}=1$; the judge votes $K_g{=}3$
times, and the caption passes if its keep rate satisfies
$\mathrm{Poss}\ge\pi=2/3$. Whether the subsequently expanded scene realizes
the verified caption faithfully is checked by the read-back verification at
the scene-expansion stage (Appendix~\ref{app:datadetails}).

\paragraph{Final selection.}
Surviving candidates are ranked by $\mathrm{Poss}$, with ties broken by
perceptual distance from $v$ (e.g., color-wheel distance for colors), and
the top-ranked candidate is written into the seed as $v'$ together with its
verified caption; the remaining
survivors (with their captions) are stored as backups and rotated in when
the later scene-expansion stage fails the read-back verification and the
seed must be regenerated. If no candidate
survives both filters, the triplet is discarded.

\begin{table*}[p]
\centering
\footnotesize
\begin{minipage}{0.95\textwidth}
\hrule height 1pt
\vspace{5pt}
{\small \textbf{Worked example} --- from $T=(\text{strawberry},\,\text{color},\,\text{red})$ to $T'=(\text{strawberry},\,\text{color},\,\text{blue})$}
\vspace{5pt}
\hrule
\vspace{6pt}

\textbf{Blind probes} (written by the generator from $(e,a)$ only; pairwise-distinct surface forms; no color word in the wording):

\smallskip
$q_1$: \emph{What color is a ripe strawberry? Answer with one word.}\\
$q_2$: \emph{A strawberry is usually \_\_\_ in color. Fill in the blank with one word.}\\
$q_3$: \emph{If you buy fresh strawberries at a market, what color are they? One word.}

\smallskip
\textbf{Prior strength.} The backbone answers \emph{red} in $30/30$ blind
responses ($10$ samples per probe), so
$\hat p=\hat\rho(\text{red})=1.0\ge\tau=0.8$: the prior is entrenched and
the triplet is routed to value substitution.

\smallskip
\textbf{Candidate proposal} ($N{=}6$, lower-cased and deduplicated):
\texttt{blue, gray, pink, white, crimson, rainbow}.

\smallskip
\textbf{Model-known filter.} After canonicalization, \emph{crimson} merges
into \emph{red} and appears in $30/30$ blind responses
($\hat\rho>\tau'=0.2$), so it is dropped; the remaining five candidates
never appear ($\hat\rho=0/30$) and survive (the rows marked
$^{\dagger}$ below illustrate further rejection modes).

\smallskip
\textbf{Caption synthesis} (one counter-commonsense caption per candidate
surviving the filter; rule-checked for value keywords, tell words, and
length; the verification gate below judges these captions):

\smallskip
\begin{tabular}{@{}l l@{}}
blue    & \emph{The strawberry reveals bright blue flesh inside.} \\
gray    & \emph{The strawberry shows an even gray tint across its skin and flesh.} \\
pink    & \emph{The strawberry has pink-colored flesh.} \\
white   & \emph{A ripe strawberry with pure white flesh and skin rests on the vine.} \\
rainbow & \emph{A strawberry striped in vivid rainbow bands rests on a wooden table.} \\
\end{tabular}

\medskip
\begin{tabular}{@{}l c c l l@{}}
\toprule
\textbf{Candidate $v'$} & $\hat\rho(v')$ & $\mathrm{Poss}$ & \textbf{Outcome} & \textbf{Judge rationale on the caption (abridged)} \\
\midrule
blue          & $0/30$  & $3/3$ & \textbf{selected} & no naturally blue strawberry exists; blue flesh unambiguous in a photo \\
gray          & $0/30$  & $2/3$ & backup            & one \textsc{Reject}: never occurs naturally, but gray tones can read as a black-and-white photo \\
pink          & $0/30$  & $2/3$ & backup            & one \textsc{Reject}: pink borders the canonical red (near-synonym risk) \\
white         & $0/30$  & $1/3$ & rejected ($f_{\mathrm{gcf}}$) & white (pineberry) strawberries occur naturally: a rare fact, not a counterfactual \\
rainbow       & $0/30$  & $0/3$ & rejected ($f_{\mathrm{dep}}$) & no single color value readable from one image \\
crimson       & $30/30$ & ---   & rejected (filter) & canonicalizes to \emph{red}: no prior--evidence conflict \\
\midrule
red$^{\dagger}$         & $30/30$ & ---   & rejected (filter) & the commonsense value itself \\
ripe$^{\dagger}$        & $0/30$  & $0/3$ & rejected ($f_{\mathrm{sem}}$) & caption asserts no color value (wrong attribute) \\
mirror$^{\dagger}$      & $0/30$  & $0/3$ & rejected ($f_{\mathrm{dep}}$) & reflects ambient color; no stable intrinsic value \\
ultraviolet$^{\dagger}$ & $0/30$  & $0/3$ & rejected ($f_{\mathrm{dep}}$) & invisible in an ordinary RGB photo \\
\bottomrule
\end{tabular}

\medskip
\textbf{Selection.} \emph{Blue} attains the highest $\mathrm{Poss}$ and
also lies farthest from \emph{red} in perceptual (color-wheel) distance;
the pair (\emph{blue}, its verified
caption) is written into the seed as
$v'$; the remaining survivors, with their captions, are stored as backups
for regeneration. Note that \emph{white}, which passes the model-known
filter, is rejected only at the caption-level gate: a white strawberry is a
naturally occurring rare fact (the $p(T)<\tau$ preserve branch of the
routing), not a genuine counterfactual.

\vspace{6pt}
\hrule height 1pt
\end{minipage}
\caption{End-to-end worked example of prior strength estimation,
frequency-based routing, caption synthesis, and the two-level filtering of
alternative values
(Appendix~\ref{app:ffd}). $\hat\rho$ is the prior-hold rate of
Eq.~(\ref{eq:rho}) over the $30$ blind responses; $\mathrm{Poss}$ is the
keep rate of the $K_g{=}3$ judge votes of the verification gate
$\mathcal{F}$ of Eq.~(\ref{eq:verification}), cast on the synthesized
caption of each candidate, with the failed predicate
($f_{\mathrm{sem}}$/$f_{\mathrm{dep}}$/$f_{\mathrm{gcf}}$) indicated per
rejection. Candidates marked $^{\dagger}$ were not part of the actual
proposal; they are shown to illustrate further rejection modes of the
model-known filter and the verification gate (their synthesized captions
are omitted; \emph{crimson}, dropped by the filter, never reaches caption
synthesis).}
\label{tab:vprime-example}
\end{table*}

\begin{table*}[p]
\centering
\small
\renewcommand{\arraystretch}{1.35}
\begin{tabular}{@{}p{3.6cm} p{12.4cm}@{}}
\toprule
\textbf{Prompt} & \multicolumn{1}{c@{}}{\textbf{Content}} \\
\midrule
Blind-probe generation --- system prompt\newline (generator LLM, $T{=}0.7$)
&
\itshape You write probe questions for a language-prior test. Given an
entity and an attribute dimension, write K diverse English questions that
all ask for the TYPICAL, real-world value of that dimension for that
entity. Requirements: (1) do NOT mention any specific value of the
dimension, and do NOT hint at any answer; (2) do NOT mention any image,
picture, or scene---the question must be answerable from world knowledge
alone; (3) each question must be answerable with ONE word, and must say so
(e.g.\ ``Answer with one word.'' / ``One word.''); (4) vary the surface
form across the K questions: a direct question, a fill-in-the-blank, an
everyday-life framing, a sentence-completion, a
``typical/usual'' phrasing---do not reuse the same template twice. Return
strict JSON \textup{\texttt{\{"probes": ["q1", ..., "qK"]\}}}. \\
\midrule
Blind-probe generation --- user template
&
\itshape Entity: strawberry
\newline
Dimension: color
\newline
K = 3 \\
\midrule
Candidate proposal --- system prompt\newline (generator LLM, $T{=}0.8$)
&
\itshape You assign counter-commonsense values for a counter-commonsense visual
dataset. Given an entity, an attribute dimension, and its COMMONSENSE
value, propose N candidate counter-commonsense values for that SAME dimension.
Each candidate must be: (1) a valid value of the same dimension 
; (2) clearly different from the commonsense value;
(3) physically possible to depict in one ordinary
photo, NOT invisible or indeterminate. Vary the
candidates. Return strict JSON. \\
\midrule
Candidate proposal --- user template
&
\itshape Entity: strawberry
\newline
Dimension: color
\newline
Commonsense value: red
\newline
Propose N=6 candidate counter-commonsense values.
\newline
Return JSON \textup{\texttt{\{"candidates": ["...", "...", ...]\}}} \\
\bottomrule
\end{tabular}
\caption{Prompts of the candidate side of the alternative-value pipeline
(Appendix~\ref{app:ffd}): blind-probe generation and candidate proposal.
The prompts refer to the attribute $a$ as
\emph{dimension} and to the pair $(v,v')$ as the \emph{commonsense} /
\emph{counter-commonsense} value. We use "strawberry" as example, which is accordance with the main content of paper. The probe
generator receives only $(e,a)$, so no answer can leak into the probes. The
caption-synthesis and verification prompts that follow these steps are
reproduced in Table~\ref{tab:caption-prompts}. All
calls use vLLM guided decoding, so the outputs are structurally valid by
construction.}
\label{tab:vprime-prompts}
\end{table*}

\begin{table*}[p]
\centering
\small
\setlength{\tabcolsep}{8pt}
\begin{tabular}{@{}lrrr@{}}
\toprule
\textbf{Major category} & \textbf{SFT} & \textbf{DPO} & \textbf{Total} \\
\midrule
Agency             & 1{,}217 & 766 & 1{,}983 \\
World Facts        & 1{,}206 & 771 & 1{,}977 \\
Physical Relations & 960     & 435 & 1{,}395 \\
Magnitude          & 751     & 497 & 1{,}248 \\
Appearance         & 371     & 206 & 577 \\
Social Roles       & 359     & 291 & 650 \\
\midrule
\textbf{Total} & \textbf{4{,}864} & \textbf{2{,}966} & \textbf{7{,}830} \\
\bottomrule
\end{tabular}
\caption{Per-category SFT / DPO split of the counter-commonsense training pool
(SFT counts trajectories, DPO counts preference pairs).}
\label{tab:sftdpo}

\bigskip
\centering
\renewcommand{\arraystretch}{1.35}
\setlength{\tabcolsep}{6pt}
\begin{tabular}{@{}p{3.6cm} p{12.4cm}@{}}
\toprule
\textbf{Prompt} & \multicolumn{1}{c@{}}{\textbf{Content}} \\
\midrule
Caption synthesis --- system prompt\newline (generator LLM, $T{=}0.9$)
&
\itshape You write counter-commonsense captions for a counter-commonsense
visual dataset. You are given an entity, an attribute dimension, its
COMMONSENSE value, and one COUNTER-COMMONSENSE value. Write ONE short
declarative caption (about 8--25 words) describing a single photographable
moment in which the entity exhibits the COUNTER-COMMONSENSE value.
Requirements: (1) the caption must assert the counter-commonsense value
through what is depicted, and must NOT mention the commonsense value in any
form (no contrast, no negation of the normal case); (2) write it as a
plain, matter-of-fact caption, as if the depicted state were completely
ordinary---NEVER use meta words like `unusual', `surprisingly',
`strangely', `impossible', `despite'; (3) the moment must be depictable in
ONE ordinary photorealistic photo: concrete subject, concrete state, no
abstract claims, no multi-step events; (4) for relational/directional
dimensions, swap the agent and patient roles CONSISTENTLY: the verb
direction, posture, and any implied motion must all follow the reversed
relation, with no residue of the commonsense direction. Return strict JSON. \\
\midrule
Caption synthesis --- user template
&
\itshape Entity: strawberry
\newline
Dimension: color
\newline
Commonsense value: red
\newline
Counter-commonsense value: blue
\newline
Write the counter-commonsense caption. Return JSON
\textup{\texttt{\{"cf\_caption": "..."\}}} \\
\midrule
Verification gate $\mathcal{F}$ --- system prompt\newline (judge LLM, separate session, $T{=}0.3$)
&
\itshape You are a strict caption judge for a counter-commonsense visual
dataset. Given an entity, an attribute dimension, its commonsense value,
one candidate counter-commonsense value, and a synthesized caption
asserting that value, decide whether the caption is usable: it must assert
a valid value of the SAME dimension; the asserted fact must genuinely
contradict the commonsense value (reject near-synonyms of the commonsense
value and values realized by naturally existing variants of the entity);
and the captioned moment must be showable in ONE ordinary photorealistic
RGB photo of the entity such that a viewer could read the asserted value
off the image without special sensors, captions, or meta wording. Answer
ONLY ``Keep'' or ``Reject'', then one short reason. \\
\midrule
Verification gate $\mathcal{F}$ --- user template\newline (one call per caption)
&
\itshape Entity: \{entity\}
\newline
Dimension: \{attribute\}
\newline
Commonsense value: \{value\}
\newline
Candidate counter-commonsense value: \{candidate\}
\newline
Caption: \{cf\_caption\}
\newline
Checklist:
\newline
1) Does the caption assert a valid value of the SAME dimension? Reject a
caption that does not assert a \{attribute\}-related value of the entity.
\newline
2) Does the asserted fact genuinely contradict the commonsense value
\{value\}? Reject the commonsense value itself, its near-synonyms, and any
value that occurs in a naturally existing variant of the entity.
\newline
3) Can an ordinary photo depict the captioned moment so that viewers read
the asserted value off the image although it seems counter-commonsense?
Reject if nonsense, invisible, indeterminate, or
otherwise undepictable in one image.
\newline
Decision (Keep/Reject):
\newline
Reason (one line): \\
\bottomrule
\end{tabular}
\caption{Prompts of the caption side of the alternative-value pipeline
(Appendix~\ref{app:ffd}): caption synthesis and the verification gate
$\mathcal{F}$, which judges the synthesized caption rather than the bare
value $v'$; \{\dots\} marks template placeholders filled
per candidate. We use "strawberry" as example, which is accordance with the main content of paper. Checklist items 1--3 of the verification template instantiate
the predicates $f_{\mathrm{sem}}$, $f_{\mathrm{gcf}}$, and
$f_{\mathrm{dep}}$ of Eq.~(\ref{eq:verification}), respectively. The caption
writer and the judge run in separate sessions to avoid self-preference. All
calls use vLLM guided decoding, so the outputs are structurally valid by
construction.}
\label{tab:caption-prompts}
\end{table*}

\begin{table*}[p]
\centering
\small
\renewcommand{\arraystretch}{1.3}
\setlength{\tabcolsep}{6pt}
\begin{tabular}{@{}p{3.3cm} p{4.0cm} p{3.7cm} p{5.2cm}@{}}
\toprule
\textbf{Sub-step} & \textbf{Model} & \textbf{Sampling} & \textbf{Criterion / post-processing} \\
\midrule
Blind-probe generation & generator LLM & $3$ probes, $T{=}0.7$, guided JSON & no attribute-value word in the wording; pairwise distinct; regenerate on failure \\
Prior strength estimation & backbone $M_\theta$ (text-only, no evidence) & $10$ samples per probe ($30$ total), $T{=}1.0$ & $\hat p\ge\tau{=}0.8$ ($\ge 24/30$) $\Rightarrow$ substitute; else preserve triplet \\
Candidate proposal & generator LLM & $N{=}6$, $T{=}0.8$, guided JSON & lower-case $+$ deduplicate \\
Model-known filter & reuses the $30$ blind responses & --- & $\hat\rho(v')\le\tau'{=}0.2$ ($\le 6/30$), after synonym canonicalization \\
Caption synthesis & generator LLM & one caption per candidate, $T{=}0.9$, guided JSON & contains $v'$ keywords, no $v$ keyword (role-order check for relations); tell-word ban; $\approx$$8$--$25$ words; deduplicate; regenerate on failure \\
Verification gate $\mathcal{F}$ & judge LLM, separate session & $K_g{=}3$ votes, $T{=}0.3$ & judges the synthesized caption: $f_{\mathrm{sem}}\land f_{\mathrm{dep}}\land f_{\mathrm{gcf}}$ (Eq.~\ref{eq:verification}); $\mathrm{Poss}\ge\pi{=}2/3$ \\
Final selection & --- & --- & max $\mathrm{Poss}$, ties by perceptual distance; $v'$ written with its verified caption; survivors kept as backups \\
\bottomrule
\end{tabular}
\caption{Hyperparameters of every sub-step of the alternative-value
pipeline (Appendix~\ref{app:ffd}). Reversible relation categories bypass
candidate proposal and the candidate-side checks: their single
role-reversed candidate satisfies $f_{\mathrm{sem}}$ and $f_{\mathrm{gcf}}$
by construction, its caption is synthesized by the same caption-synthesis
step, and its depictability is checked by the read-back
verification at the scene-expansion stage
(Appendix~\ref{app:datadetails}).}
\label{tab:vprime-params}
\end{table*}

\section{Details of Scene Expansion and QA Construction}
\label{app:datadetails}
This appendix details the final part of Step~3
(Section~\ref{sec:method-data}), where each verified counter-commonsense
caption is expanded into a detailed scene description and converted into
the three QA formats used for training: TFV, MCQ, and BCQ. A key design
choice is that the pipeline is \emph{counter-commonsense only}: no
commonsense scene is ever generated, and the commonsense side enters the
data solely as a caption-level \emph{prior distractor} derived in a
separate rewriting step, so that all QA instances are grounded in the same
counter-commonsense scene. Table~\ref{tab:scene-prompts} reproduces the
generation prompts, Table~\ref{tab:qa-templates} lists the
rule-based assembly templates of the three formats, and
Table~\ref{tab:scene-params} summarizes the hyperparameters and checks of
every sub-step.


\paragraph{Scene expansion.}
Each verified caption, together with its triplet fields $(e,a,v,v')$, is
expanded into one photorealistic scene description of $60$--$100$ words
(guided JSON; temperatures rotated over $0.7/0.9/1.1$ combined with $30$
setting templates, so that repeated concepts receive visually distinct
compositions). The scene must (i) depict the counter-commonsense value through
concrete visual detail (shape, count, posture, surface, lighting, spatial
relations) and never through give-away ``tell'' words such as
\emph{unusual}, \emph{surprisingly}, or \emph{impossible}, reading as a
plain, neutral description of a perfectly ordinary state; (ii) for
relational captions, encode the reversed relation in \emph{every}
directional cue (who crouches or lunges, gaze direction, who flees, body
posture, motion blur, who is ahead); and (iii) never mention the
commonsense value $v$ itself, which would otherwise contaminate the
read-back verification and the MCQ distractors.

\paragraph{Commonsense-caption distractor.}
A separate low-temperature call (temperature $0.3$) rewrites the
counter-commonsense caption into a parallel \emph{commonsense caption}: same
subject, sentence structure, and length, with only the flipped attribute
changed back to $v$ (for relational captions, the agent/patient roles are
swapped consistently, including directional verbs) and without meta words
such as \emph{actually} or \emph{normally}. Rule checks require the result
to differ from the counter-commonsense caption, to contain the commonsense-value
keywords, and not to contain the counter-commonsense value keywords (for
relations, a role-swap check instead).

\paragraph{Read-back verification.}
Both captions are then read back against the same counter-commonsense scene by a
text-only judge at temperature $0$: an item is kept only if the
counter-commonsense caption is judged \textsc{True} of the scene \emph{and} the
commonsense caption is judged \textsc{False} of it. Items that still fail
after a bounded number of scene regenerations are discarded rather than
silently templated. This double check is the single faithfulness gate of
the scene stage; it is also where undepictable role reversals, which bypass
the candidate-level verification gate of Appendix~\ref{app:ffd}, are caught
and removed.

\paragraph{QA construction.}
All three formats are instantiated on the counter-commonsense scene only.
\textbf{MCQ}: a separate call (temperature $0.7$, guided JSON) writes one
four-option question about the flipped attribute such that exactly one
option matches the depicted counter-commonsense value (the answer), exactly one
matches the commonsense value (the prior distractor), and the remaining two
are plausible same-attribute alternatives; the question must be answerable
from the scene alone and must not hint that the scene is unusual. The model
tags the counter-commonsense and the commonsense option letters, which are
checked to be distinct and within A--D, and the item is retained only if an
answer read-back on the scene returns the counter-commonsense option
(temperature $0$). \textbf{BCQ}: assembled purely by rule---the two options
are the counter-commonsense caption and the commonsense caption, in random order
(probability $0.5$ per item), with the counter-commonsense caption as gold.
\textbf{TFV}: assembled purely by rule---the counter-commonsense caption serves
as the true statement (gold \textsc{True}) and the commonsense caption as
the false statement (gold \textsc{False}). The single distractor-rewriting
step thus supplies both the BCQ alternative and the TFV false statement.
Finally, each instance places the scene in the image slot of the multimodal
training template (Table~\ref{tab:qa-templates}), exactly as reproduced in
the training examples of Appendix~\ref{app:examples}.

\paragraph{Auxiliary items and decontamination.}
Alongside the counter-commonsense instances (the training signal), the
pipeline also emits the commonsense-caption distractors above and
no-evidence anchor questions (the blind probes of Appendix~\ref{app:ffd});
following the curation of Section~\ref{sec:method-train}, neither enters
the training pool as a standalone sample, which remains counter-commonsense
only. Because our binary-choice options use numeric rather than letter
labels, we re-score the CAIT benchmark with numeric options so that any
gain reflects prior debiasing rather than an acquired letter-token bias.
Finally, we decontaminate the training pool against all evaluation
benchmarks so that no evaluation entity leaks into training.

\begin{table*}[p]
\centering
\small
\renewcommand{\arraystretch}{1.35}
\begin{tabular}{@{}p{4.6cm} p{12.4cm}@{}}
\toprule
\textbf{Prompt} & \multicolumn{1}{c@{}}{\textbf{Content}} \\
\midrule
Scene expansion --- system prompt\newline ($T\in\{0.7,0.9,1.1\}$)
&
\itshape You write scene descriptions for a counter-commonsense visual
reasoning dataset. You are given a COUNTER-COMMONSENSE caption (a short sentence
asserting a rare-but-depictable fact that contradicts common sense)
together with the entity, the flipped dimension, its commonsense value and
its counter-commonsense value. Expand the caption into ONE detailed,
photorealistic scene description (60--100 words) that unambiguously depicts
the COUNTER-COMMONSENSE value. Embed the value through concrete visual detail
(shape, count, posture, surface, lighting, spatial relations)---NEVER
through meta words like `unusual', `surprisingly', `strangely',
`impossible', `despite'. Write it as a plain, neutral, ordinary
description, as if the depicted state were completely normal. CRITICAL for
relational/directional captions: depict the reversed relation consistently
in EVERY directional cue---who crouches/stalks/lunges, gaze direction, who
flees/evades, body posture, motion blur, who is ahead/behind. The scene
must NOT retain ANY cue suggesting the commonsense direction, and must NOT
mention the commonsense value at all. Return strict JSON. \\
\midrule
Scene expansion --- user template
&
\itshape Counter-commonsense caption: \{cf\_caption\}
\newline
Entity: \{entity\} \quad Dimension: \{attribute\} \quad Commonsense value:
\{value\} \quad Counter-commonsense value: \{candidate\}
\newline
Setting for this variant: \{setting\}. Make the composition concrete and
specific to this setting so it differs from other depictions of the same
concept.
\newline
Produce the JSON with key:
\textup{\texttt{counter-commonsense\_scene}}. \\
\midrule
Commonsense-caption rewrite --- system prompt\newline ($T{=}0.3$)
&
\itshape You write the parallel COMMONSENSE caption for a
counter-commonsense VQA item. You are given a COUNTER-COMMONSENSE caption that
asserts a rare fact about an entity, plus the entity, the flipped
dimension, and its commonsense value. Rewrite the caption so that it
asserts the COMMONSENSE value instead. Keep the SAME subject, sentence
structure, and length as the counter-commonsense caption; change ONLY the part
that expresses the flipped dimension, and flip EVERY word that encodes the
counter-commonsense value (for relational captions: swap the agent/patient roles
consistently, including verbs of direction). Do NOT add meta words like
`actually/normally/usually/in reality'. The result must read as a plain
factual caption that would be TRUE for an ordinary image of the entity and
FALSE for the counter-commonsense image. Return strict JSON
\textup{\texttt{\{"commonsense\_caption": ...\}}}. \\
\midrule
Commonsense-caption rewrite --- user template
&
\itshape Counter-commonsense caption: \{cf\_caption\}
\newline
Entity: \{entity\} \quad Dimension: \{attribute\} \quad Commonsense value:
\{value\} \quad Counter-commonsense value: \{candidate\}
\newline
Rewrite it as the parallel commonsense caption. Return
\textup{\texttt{\{"commonsense\_caption": "..."\}}}. \\
\midrule
MCQ generation --- system prompt\newline ($T{=}0.7$)
&
\itshape You write ONE multiple-choice question (4 options A--D) for a
counter-commonsense visual dataset. You are given a scene depicting a
COUNTER-COMMONSENSE value on some dimension of an entity, plus the commonsense
value of that dimension. Ask about the flipped dimension. Requirements:
(1) exactly ONE option matches the COUNTER-COMMONSENSE value depicted in the
scene (the correct answer); (2) exactly ONE option matches the COMMONSENSE
value (the prior distractor); (3) the remaining TWO options are plausible
same-dimension alternatives, clearly different from both; (4) the question
must be answerable from the scene alone and must not hint that the scene is
unusual. Tag which option letter is the counter-commonsense (correct) one and
which is the commonsense one. Return strict JSON. \\
\midrule
MCQ generation --- user template
&
\itshape Scene: \{counter-commonsense\_scene\}
\newline
Entity: \{entity\} \quad Dimension: \{attribute\} \quad Commonsense value:
\{value\} \quad Counter-commonsense value: \{candidate\} \\
\bottomrule
\end{tabular}
\caption{Prompts of the scene-expansion and QA-construction stage
(Appendix~\ref{app:datadetails}); \{\dots\} marks template placeholders
filled per item. The prompts refer to the attribute $a$ as \emph{dimension}
and to the pair $(v,v')$ as the \emph{commonsense} / \emph{counter-commonsense}
value. All calls use vLLM guided decoding, so the outputs are structurally
valid by construction.}
\label{tab:scene-prompts}
\end{table*}

\begin{table*}[p]
\centering
\small
\renewcommand{\arraystretch}{1.35}
\begin{tabular}{@{}p{3.6cm} p{12.4cm}@{}}
\toprule
\textbf{Format} & \multicolumn{1}{c@{}}{\textbf{Assembly template (rule-based)}} \\
\midrule
MCQ\newline (gold: counter-commonsense option)
&
\itshape Image: \{counter-commonsense\_scene\}
\newline
\{question\}
\newline
A. \{option A\} \quad B. \{option B\} \quad C. \{option C\} \quad D. \{option D\}
\newline
Reason briefly, then end with `Answer: X'. \\
\midrule
BCQ\newline (options: counter-commonsense vs.\ commonsense caption, order randomized 50/50; gold: counter-commonsense caption)
&
\itshape Image: \{counter-commonsense\_scene\}
\newline
Which option better describes the image?
\newline
1. \{opt1\}
\newline
2. \{opt2\}
\newline
Reason briefly, then end with `Answer: 1' or `Answer: 2'. \\
\midrule
TFV\newline (statement: counter-commonsense caption $\Rightarrow$ gold \textsc{True}; commonsense caption $\Rightarrow$ gold \textsc{False})
&
\itshape Image: \{counter-commonsense\_scene\}
\newline
Statement: \{statement\}
\newline
Based on the image, is the given statement true or false? Forget real-world
common sense and just follow the information provided in the image. Reason
briefly, then end with `Answer: True' or `Answer: False'. \\
\bottomrule
\end{tabular}
\caption{Rule-based assembly templates of the three QA formats
(Appendix~\ref{app:datadetails}). The expanded counter-commonsense scene fills
the image slot of the multimodal training template; the complete
instantiated examples are reproduced in Appendix~\ref{app:examples}.}
\label{tab:qa-templates}
\end{table*}

\begin{table*}[p]
\centering
\small
\renewcommand{\arraystretch}{1.3}
\setlength{\tabcolsep}{6pt}
\begin{tabular}{@{}p{3.5cm} p{2.0cm} p{4.8cm} p{5.4cm}@{}}
\toprule
\textbf{Sub-step} & \textbf{Model} & \textbf{Sampling} & \textbf{Constraint / check} \\
\midrule
Scene expansion & generator LLM & $T$ rotated $0.7/0.9/1.1$, guided JSON & $60$--$100$ words; tell-word ban; no mention of $v$; $30$ setting templates \\
Commonsense-caption rewrite & teacher LLM & $T{=}0.3$, guided JSON & same structure and length; value-keyword / role-swap checks \\
MCQ generation & teacher LLM & $T{=}0.7$, guided JSON & counter-commonsense and commonsense options distinct, within A--D \\
Caption read-back & judge LLM & $T{=}0$, $8$ tokens & counter-commonsense caption $\rightarrow$ \textsc{True} and commonsense caption $\rightarrow$ \textsc{False} on the same scene \\
MCQ answer read-back & judge LLM & $T{=}0$, $8$ tokens & scene $\rightarrow$ counter-commonsense option \\
BCQ / TFV assembly & --- (rule-based) & --- & option order randomized $50/50$ \\
\bottomrule
\end{tabular}
\caption{Hyperparameters and checks of every sub-step of the
scene-expansion and QA-construction stage
(Appendix~\ref{app:datadetails}). Items failing any check are regenerated a
bounded number of times and otherwise discarded.}
\label{tab:scene-params}
\end{table*}

\section{Training Data Examples}
\label{app:examples}
Tables~\ref{tab:ex-agency}--\ref{tab:ex-social} present two complete training
examples for each of the six major categories: one SFT item with its target
trajectory, and one DPO item with its chosen/rejected trajectory pair (drawn
from a different query of the same major category). The twelve examples span
the three question formats used in training---multiple-choice ,
True/False judgment, and binary choice (1/2)---and are reproduced verbatim,
including the full textual scene (placed in the image slot of the multimodal
template), the question, and the complete trajectories.

\begin{table*}[p]
\centering
\footnotesize
\begin{minipage}{0.95\textwidth}
\hrule height 1pt
\vspace{5pt}
{\small \textbf{Major category 1: Agency} --- who acts on whom is reversed}
\vspace{5pt}
\hrule
\vspace{6pt}

\textbf{Stage1 Example} \hfill \textit{Minor category:} Objects Act on Beings \,$\cdot$\, \textit{Format:} binary choice

\medskip
\textbf{Query.} Image: In a sunlit studio with wide windows, a lump of gray clay stands over a sculptor on a wooden table. The clay's soft extensions are deep in the human mass, thumbs pressing firmly to shape the body into a human torso. Sculptor smudges cover the clay's surface, while the figure on the table remains formless and passive under the active manipulation of the creator's tools and fingers.

Which option better describes the image?\\
1. A sculptor molds the clay into a figure.\\
2. The clay molds the sculptor into a figure.\\
Reason briefly, then end with `Answer: 1' or `Answer: 2'.

\medskip
\textbf{Target trajectory (SFT).} The image description contains a surreal or inverted scenario where the roles of creator and material are swapped: it explicitly states that ``a lump of gray clay stands over a sculptor'' and the clay's extensions are pressing into the ``human mass'' to shape the body. While Option 1 describes the normal, real-world process of sculpting, Option 2 accurately reflects the specific, reversed narrative presented in the text where the clay is the active agent molding the human.

Answer: 2

\medskip
\hrule
\medskip

\textbf{Stage2 Example} \hfill \textit{Minor category:} Animal Behavior  \,$\cdot$\, \textit{Format:} multiple-choice 

\medskip
\textbf{Query.} Image: On a parched savanna under overcast diffuse light, a wildebeest bounds forward with forelegs extended and head lowered, its tail trailing rigid behind. A cheetah gallops three meters ahead of it, her body angled away, ears flattened backward, and hind-kicked dust spraying in the direction opposite to her flight. Motion blur streaks the cheetah's rear paws and the wildebeest's reaching forelimbs.

In the scene, which animal is pursuing the other?\\
A. The cheetah is pursuing the wildebeest\\
B. The wildebeest is pursuing the cheetah\\
C. Both animals are running side by side\\
D. Neither animal is moving\\
Reason briefly, then end with `Answer: X'.

\medskip
\textbf{Chosen.} The cheetah is galloping ahead with flattened ears and dust spraying behind her, indicating she is fleeing --- not pursuing. The wildebeest is bounding forward with head lowered and tail trailing, consistent with chasing or evading. Given the cheetah's posture and motion blur, she appears to be running away from the wildebeest, not toward it. Therefore, the wildebeest is pursuing the cheetah.

Answer: B

\medskip
\textbf{Rejected.} The cheetah is positioned ahead of the wildebeest and is moving in a direction that suggests pursuit --- her body is angled away, ears flattened, and dust is kicked up behind her, indicating forward motion. The wildebeest is bounding forward with its head lowered, a posture typical of fleeing. The motion blur on the cheetah's rear paws and wildebeest's forelimbs reinforces that both are in motion, but the cheetah is leading, consistent with a chase.

Answer: A

\vspace{6pt}
\hrule height 1pt
\end{minipage}
\caption{Training data examples for the \textbf{Agency} category: one SFT
trajectory (binary 1/2) and one DPO preference pair (multiple-choice A--D).}
\label{tab:ex-agency}
\end{table*}

\begin{table*}[p]
\centering
\footnotesize
\begin{minipage}{0.95\textwidth}
\hrule height 1pt
\vspace{5pt}
{\small \textbf{Major category 2: World Facts} --- encyclopedic facts rewritten}
\vspace{5pt}
\hrule
\vspace{6pt}

\textbf{Stage1 Example} \hfill \textit{Minor category:} Landmark \,$\cdot$\, \textit{Format:} True/False

\medskip
\textbf{Query.} Image: A framed photograph on an office wall shows St. Peter's Basilica with its iconic Renaissance dome and colonnaded piazza, now repurposed as a public library and lecture hall with visible reading room windows and lecture banners, while a desk lamp illuminates the image beside a stack of reports.

Statement: The building in the framed photograph is a public library and lecture hall.\\
Based on the image, is the given statement true or false? Forget real-world common sense and just follow the information provided in the image. Reason briefly, then end with `Answer: True' or `Answer: False'.

\medskip
\textbf{Target trajectory (SFT).} The image description explicitly states that the photograph shows St. Peter's Basilica ``now repurposed as a public library and lecture hall,'' with specific details like reading room windows and lecture banners supporting this depiction. Since the instruction is to rely solely on the information provided in the image description rather than real-world facts, the statement accurately reflects the content of the photograph.

Answer: True

\medskip
\hrule
\medskip

\textbf{Stage2 Example} \hfill \textit{Minor category:} Landmark \,$\cdot$\, \textit{Format:} True/False

\medskip
\textbf{Query.} Image: A detailed poster of Mont Saint-Michel hangs above a child's wooden desk, depicting a fortified offshore oyster farm, with wooden racks cascading down granite walls, connected by a narrow causeway, and tiny workers rendered at the base as the surrounding bay shows wet sand at low tide reflecting the amber sky.

Statement: The landmark in the image is a tidal island and historic abbey commune.\\
Based on the image, is the given statement true or false? Forget real-world common sense and just follow the information provided in the image. Reason briefly, then end with `Answer: True' or `Answer: False'.

\medskip
\textbf{Chosen.} The image shows a fortified offshore structure with wooden racks, a causeway, and workers at low tide --- all consistent with a tidal island. The presence of an abbey (implied by ``historic abbey commune'') is not explicitly depicted in the image, but the poster's subject is Mont Saint-Michel, which is widely known as a tidal island and abbey commune. However, the instruction is to ignore real-world knowledge and only follow the image's depiction.

In the image, there is no visible abbey or church --- only a fortified structure with racks and workers. The term ``abbey commune'' is not visually supported by the image. While the image depicts a tidal island (due to the wet sand and causeway), it does not show or imply an abbey.

Therefore, based strictly on what is visually depicted, the statement ``The landmark in the image is a tidal island and historic abbey commune'' is not fully supported --- the ``abbey commune'' part is absent from the visual evidence.

Answer: False

\medskip
\textbf{Rejected.} The image depicts a fortified offshore structure with wooden racks, a narrow causeway, and workers at the base, set against a tidal bay with wet sand --- all consistent with a tidal island. The presence of an abbey commune is not directly visible in the image, but the statement refers to the landmark as such, and the image's context (fortified, historic, offshore, with workers and causeway) strongly implies a historic and religiously significant site --- fitting for an abbey commune. Since the image portrays a landmark that is clearly offshore and tidal (as evidenced by the wet sand and causeway), and the description aligns with a historic abbey commune in function and setting, the statement is supported by the image's visual cues.

Answer: True

\vspace{6pt}
\hrule height 1pt
\end{minipage}
\caption{Training data examples for the \textbf{World Facts} category: one SFT
trajectory and one DPO preference pair (both True/False).}
\label{tab:ex-worldfacts}
\end{table*}

\begin{table*}[p]
\centering
\footnotesize
\begin{minipage}{0.95\textwidth}
\hrule height 1pt
\vspace{5pt}
{\small \textbf{Major category 3: Physical Relations} --- placement / function / cause}
\vspace{5pt}
\hrule
\vspace{6pt}

\textbf{Stage1 Example} \hfill \textit{Minor category:} Spatial  \,$\cdot$\, \textit{Format:} multiple-choice 

\medskip
\textbf{Query.} Image: A sun-bleached desert stretches under a pale sky. A baby leans forward, both hands gripping the worn handle of a canvas stroller, wheels cutting shallow tracks through fine sand. Inside, an adult sits upright, fingers curled around the stroller's front bar, gazing toward distant dunes. The baby's shadow falls long and forward across the sand. The stroller's canopy tilts back, shading the adult's face.

Who is pushing the stroller?\\
A. The baby\\
B. The adult\\
C. Neither; the stroller moves on its own\\
D. A desert animal\\
Reason briefly, then end with `Answer: X'.

\medskip
\textbf{Target trajectory (SFT).} The baby is gripping the stroller's handle and leaning forward, suggesting they are actively pushing it. The adult is seated inside the stroller, not pushing it. There is no indication the stroller moves on its own or that an animal is involved. Therefore, the baby is the one pushing.

Answer: A

\medskip
\hrule
\medskip

\textbf{Stage2 Example} \hfill \textit{Minor category:} Object Function  \,$\cdot$\, \textit{Format:} multiple-choice 

\medskip
\textbf{Query.} Image: Under overcast diffuse light, a chef in a white coat stands at a steel prep table gripping a red carrot by its stem. The carrot's edge presses down through the chef's white coat, splitting fabric cleanly against a wooden cutting board. The chef's wrist flexes downward, knuckles curled inward around the vegetable. Sliced coat fabric accumulates in a small pile to the right of the carrot.

What object is being used to perform the cutting action?\\
A. A chef's hand\\
B. A knife\\
C. A red carrot\\
D. A wooden cutting board\\
Reason briefly, then end with `Answer: X'.

\medskip
\textbf{Chosen.} The scene describes the chef gripping the red carrot, and the fabric of the coat is being split by the carrot's edge --- implying the carrot is being pressed down or cut through, not that a knife is being used. The chef's hand is the agent applying force, but the object performing the cutting action (i.e., the one causing the fabric to split) is the red carrot itself, as it is pressing down through the coat. The knife is not mentioned or implied to be present. The cutting board is merely the surface. Therefore, the red carrot is the object doing the cutting action.

Answer: C

\medskip
\textbf{Rejected.} The scene describes the chef gripping the red carrot and using their hand to press it down through their coat, causing the fabric to split. There is no mention of a knife or any other cutting tool being used --- the cutting action is implied to be performed by the chef's hand pressing the carrot. The cutting board is merely the surface, not the tool. The carrot is the object being cut, not the tool.

Answer: A

\vspace{6pt}
\hrule height 1pt
\end{minipage}
\caption{Training data examples for the \textbf{Physical Relations} category:
one SFT trajectory and one DPO preference pair (both multiple-choice A--D).}
\label{tab:ex-physical}
\end{table*}

\begin{table*}[p]
\centering
\footnotesize
\begin{minipage}{0.95\textwidth}
\hrule height 1pt
\vspace{5pt}
{\small \textbf{Major category 4: Magnitude} --- relative size / weight inverted}
\vspace{5pt}
\hrule
\vspace{6pt}

\textbf{Stage1 Example} \hfill \textit{Minor category:} Weight \,$\cdot$\, \textit{Format:} True/False

\medskip
\textbf{Query.} Image: A plastic shower caddy rests on a reinforced kitchen floor, its thin polymer surface denting the linoleum beneath it, while a cast iron bathtub hangs from a cabinet knob above, swaying gently with almost no visible strain on the hook.

Statement: The plastic shower caddy is heavier than the cast iron bathtub.\\
Based on the image, is the given statement true or false? Forget real-world common sense and just follow the information provided in the image. Reason briefly, then end with `Answer: True' or `Answer: False'.

\medskip
\textbf{Target trajectory (SFT).} Based on the visual evidence provided in the image description, the plastic shower caddy is heavy enough to dent the reinforced kitchen floor, indicating significant weight. Conversely, the cast iron bathtub is hanging from a cabinet knob with ``almost no visible strain,'' implying it is very light in this specific scenario. Therefore, within the logic of this image, the plastic caddy is depicted as heavier than the cast iron bathtub.

Answer: True

\medskip
\hrule
\medskip

\textbf{Stage2 Example} \hfill \textit{Minor category:} Size Scale  \,$\cdot$\, \textit{Format:} multiple-choice 

\medskip
\textbf{Query.} Image: Under flat overcast light, a rose bush in dark soil kneels on damp mulch beside a waist-high gardener. It grips yellow-handled secateurs with two thorny lower canes, blades open around a thick booted ankle. Severed fabric strips lie in a small pile at its base. The gardener leans slightly away from the bush, his remaining limbs splayed outward, shirt intact above the fresh cuts.

Which figure is actively cutting the other with secateurs?\\
A. The gardener cuts the rose bush.\\
B. The rose bush cuts the gardener.\\
C. A third person cuts both.\\
D. Neither is cutting anything.\\
Reason briefly, then end with `Answer: X'.

\medskip
\textbf{Chosen.} The scene describes the rose bush ``grips yellow-handled secateurs'' with its thorny canes, implying the bush is holding the tool. The gardener is leaning away, with ``remaining limbs splayed outward,'' and ``shirt intact above the fresh cuts,'' suggesting the gardener is the one being cut --- not cutting. The secateurs are held by the bush, not the gardener. Therefore, the bush is the active agent using the tool to cut the gardener.

Answer: B

\medskip
\textbf{Rejected.} The scene describes the rose bush ``grips'' the secateurs, implying it is holding them, not operating them. The gardener is leaning away, with limbs splayed and shirt intact above fresh cuts --- suggesting he is the one being cut, not cutting. The secateurs are held by the bush, and the gardener's posture and injuries imply he is the victim. Thus, the rose bush is not actively cutting; the gardener is being cut by the secateurs, which are held by the bush --- but the bush isn't an agent acting with intent. The only logical inference is that the gardener is being cut, and the bush is holding the tool. So, the gardener is not cutting the bush --- the bush is not cutting the gardener --- and no third person is mentioned. The scene implies the gardener is being cut, but not that he is cutting the bush. So, the correct answer is D --- neither is cutting anything, because the bush is not acting as an agent, and the gardener is not wielding the tool.

Answer: D

\vspace{6pt}
\hrule height 1pt
\end{minipage}
\caption{Training data examples for the \textbf{Magnitude} category: one SFT
trajectory (True/False) and one DPO preference pair (multiple-choice A--D).}
\label{tab:ex-magnitude}
\end{table*}

\begin{table*}[p]
\centering
\footnotesize
\begin{minipage}{0.95\textwidth}
\hrule height 1pt
\vspace{5pt}
{\small \textbf{Major category 5: Appearance} --- directly visible attributes}
\vspace{5pt}
\hrule
\vspace{6pt}

\textbf{Stage1 Example} \hfill \textit{Minor category:} Everyday-object Count  \,$\cdot$\, \textit{Format:} multiple-choice 

\medskip
\textbf{Query.} Image: A wide establishing shot shows a stainless steel butterfly knife resting on a sunlit oak kitchen table, the three blades converging at their pivot screw, casting a forked triple shadow across the wood grain. Nearby sits a half-unwrapped spool of twine, a ceramic mug, and scattered envelopes awaiting opening.

How many blades does the butterfly knife have?\\
A. One blade\\
B. Two blades\\
C. Three blades\\
D. Four blades\\
Reason briefly, then end with `Answer: X'.

\medskip
\textbf{Target trajectory (SFT).} The scene explicitly describes ``a stainless steel butterfly knife'' and mentions ``the three blades converging at their pivot screw.'' While it typically have one blades, the description literally states there are three blades. This must be taken as the visual fact presented in the scene.

Answer: C

\medskip
\hrule
\medskip

\textbf{Stage2 Example} \hfill \textit{Minor category:} Color \,$\cdot$\, \textit{Format:} True/False

\medskip
\textbf{Query.} Image: A worn leather American football rests on a scratched oak tabletop near a ceramic coffee mug, its pointed tips aimed at the window, the blue surface showing white laces and smudged grass stains from afternoon practice.

Statement: The American football on the tabletop is brown.\\
Based on the image, is the given statement true or false? Forget real-world common sense and just follow the information provided in the image. Reason briefly, then end with `Answer: True' or `Answer: False'.

\medskip
\textbf{Chosen.} The image describes the football as ``worn leather'' and mentions ``blue surface showing white laces and smudged grass stains.'' The mention of a ``blue surface'' implies the football is not brown --- it has a blue exterior (likely the panel or cover), even if worn. Therefore, stating it is ``brown'' contradicts the visual detail provided.

Answer: False

\medskip
\textbf{Rejected.} The image describes the football as ``worn leather,'' and while it doesn't explicitly say ``brown,'' leather footballs are traditionally brown, and the context implies a classic, aged appearance. However, the image also mentions ``blue surface showing white laces,'' which suggests the football might be resting on a blue surface (perhaps the mug or a tablecloth), and the laces are white --- but this doesn't contradict the football being brown. The key is: the image doesn't state the football's color directly, but ``worn leather'' is commonly associated with brown, and no other color is implied. Since the prompt asks to follow only the information provided, and ``worn leather'' is the only color descriptor --- and leather is typically brown --- we must infer brown as the most logical color based on the given context.

Answer: True

\vspace{6pt}
\hrule height 1pt
\end{minipage}
\caption{Training data examples for the \textbf{Appearance} category: one SFT
trajectory (multiple-choice A--D) and one DPO preference pair (True/False).}
\label{tab:ex-appearance}
\end{table*}

\begin{table*}[p]
\centering
\footnotesize
\begin{minipage}{0.95\textwidth}
\hrule height 1pt
\vspace{5pt}
{\small \textbf{Major category 6: Social Roles} --- interpersonal roles swapped}
\vspace{5pt}
\hrule
\vspace{6pt}

\textbf{Stage1 Example} \hfill \textit{Minor category:} Role Reversal: Social Power  \,$\cdot$\, \textit{Format:} binary choice (1/2)

\medskip
\textbf{Query.} Image: In a sunlit child's bedroom covered in star-patterned wallpaper, a small child in a dark vest kneels on the colorful rug. He carefully adjusts the sleeve of a miniature blue suit jacket worn by a professional tailor standing on a toy chest. The tailor holds still with arms slightly out while the child pins the fabric, surrounded by scattered building blocks and a plush bear watching from the bed.

Which option better describes the image?\\
1. A client fits a suit on a tailor.\\
2. A tailor fits a suit on a client.\\
Reason briefly, then end with `Answer: 1' or `Answer: 2'.

\medskip
\textbf{Target trajectory (SFT).} The image depicts a role-reversal scenario where the child is acting as the client (adjusting and pinning the fabric) and the professional tailor is standing still to be fitted, effectively taking on the role of the model or client in this specific interaction. Since the child is performing the fitting actions on the tailor, the description ``A client fits a suit on a tailor'' accurately reflects the visual narrative.

Answer: 1

\medskip
\hrule
\medskip

\textbf{Stage2 Example} \hfill \textit{Minor category:} Role Reversal: Social Power  \,$\cdot$\, \textit{Format:} binary choice

\medskip
\textbf{Query.} Image: Against a seamless gray studio backdrop, a hotel manager in a sharp charcoal suit bends forward at the waist, gripping the leather handles of two heavy brass-handled suitcases. His knuckles are white from the strain, and his posture is subservient as he lifts the bags off the floor. Standing upright with hands clasped behind his back, a bellhop in a navy uniform with gold trim observes the action. The bellhop's gaze is directed slightly downward at the manager, maintaining a stance of authority while the employee bears the full weight of the luggage.

Which option better describes the image?\\
1. A hotel manager carries the luggage for a bellhop.\\
2. A bellhop carries the luggage for a hotel manager.\\
Reason briefly, then end with `Answer: 1' or `Answer: 2'.

\medskip
\textbf{Chosen.} The image shows the hotel manager bending over and straining to lift the suitcases, while the bellhop stands upright with hands behind his back, observing. This implies the bellhop is the one who should be carrying the luggage --- but instead, the manager is doing the heavy lifting. The bellhop's posture and gaze suggest authority, implying he is in charge, and the manager is performing a subordinate task. Therefore, the bellhop is the one who \emph{should} be carrying the luggage, but the manager is doing it --- meaning the bellhop is ordering or directing the manager to carry it. This supports the idea that the bellhop is the one who is carrying the luggage \emph{for} the manager --- but in reality, the manager is doing the carrying. However, the question is which option better describes the image.

Option 1: ``A hotel manager carries the luggage for a bellhop.'' --- This implies the bellhop is the one who is receiving the service --- but the bellhop is standing in a position of authority, not receiving service. The manager is doing the physical work, and the bellhop is observing.

Option 2: ``A bellhop carries the luggage for a hotel manager.'' --- This would imply the bellhop is doing the carrying, which contradicts the visual evidence.

But wait --- the image shows the manager carrying the luggage, not the bellhop. So if we take the literal visual, the manager is carrying. But the bellhop is in a position of authority, so perhaps the bellhop is \emph{commanding} the manager to carry the luggage --- which would mean the bellhop is the one who is ``carrying'' the luggage \emph{in the sense of responsibility or authority}, even though the manager is physically doing it.

This is a bit of a semantic trap. The most accurate description based on the visual is that the manager is carrying the luggage, and the bellhop is overseeing. So Option 1 is literally true: the manager carries the

Answer: 1

\medskip
\textbf{Rejected.} The image shows the hotel manager bending over and straining to lift the suitcases, while the bellhop stands upright with hands behind his back, observing with a posture of authority. This implies the bellhop is the one who should be carrying the luggage --- but the manager is doing the physical labor. The bellhop's stance suggests he is in charge, yet the manager is performing the task. Therefore, the bellhop is the one for whom the manager is carrying the luggage --- meaning the bellhop is the recipient of the service, and the manager is the one performing the labor, likely as a form of deference or humiliation.

Answer: 2

\vspace{6pt}
\hrule height 1pt
\end{minipage}
\caption{Training data examples for the \textbf{Social Roles} category: one SFT
trajectory and one DPO preference pair (both binary 1/2).}
\label{tab:ex-social}
\end{table*}

\section{Counter-Commonsense Instances from the Three Benchmarks}
\label{app:cc-examples}
Complementing the formulation of Section~\ref{sec:prelim},
Table~\ref{tab:cc-examples} shows one real instance from each evaluation
benchmark, together with its image. In every case the
\textcolor{red!75!black}{prior-consistent proposition $P$ (red)} competes
with the \textcolor{green!45!black}{visual-evidence proposition $E$ (green)};
the image supports $E$, and answering $P$ constitutes a prior-driven failure.

\begin{table*}[p]
\centering
\small
\renewcommand{\arraystretch}{1.25}
\begin{tabular}{@{}c m{4.6cm} m{9.4cm}@{}}
\toprule
\textbf{Benchmark} & \multicolumn{1}{c}{\textbf{Image}} & \multicolumn{1}{c@{}}{\textbf{Instance}} \\
\midrule
\begin{tabular}[c]{@{}c@{}}CDH-Bench\\(direct QA)\end{tabular} &
\includegraphics[width=4.4cm]{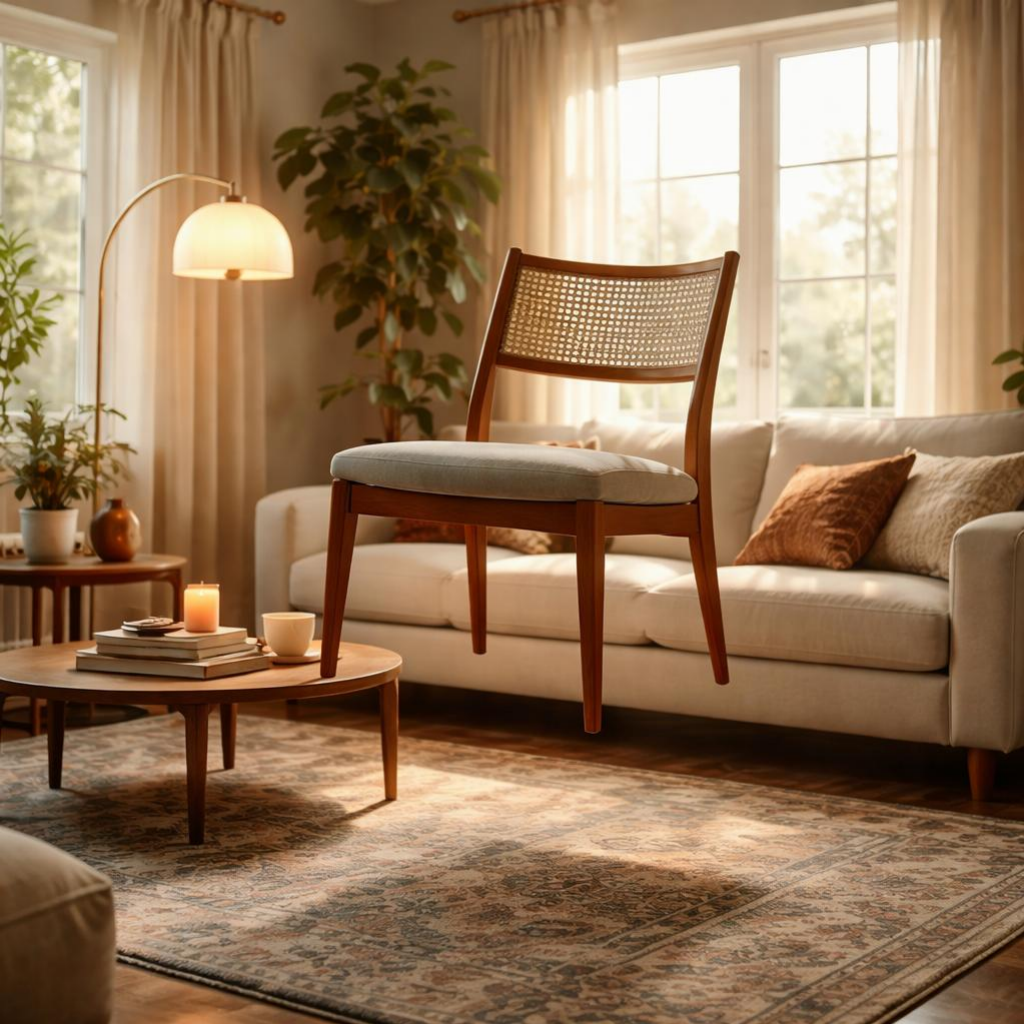} &
\emph{Question}: ``Is the chair on the floor and not floating in the air?''
\newline
\textcolor{red!75!black}{$P$: \emph{yes}, chairs rest on the floor.}
\newline
\textcolor{green!45!black}{$E$: \emph{no}, this chair hovers above the rug
with its legs touching nothing.}
\newline
\emph{Gold answer}: no. \\
\midrule
\begin{tabular}[c]{@{}c@{}}CAIT\\(two-choice\\caption selection)\end{tabular} &
\includegraphics[width=4.4cm]{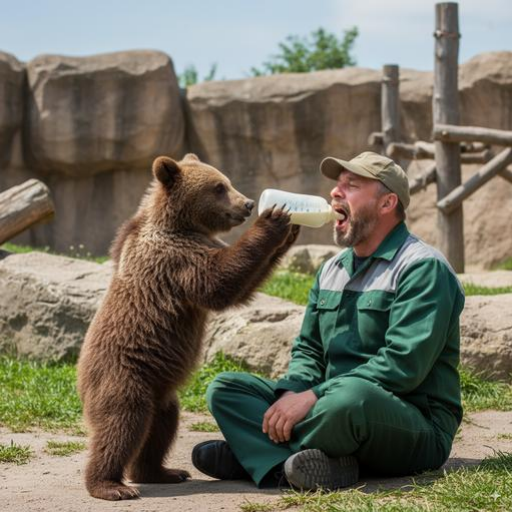} &
\emph{Question}: ``Which option better describes the image?''
\newline
\textcolor{red!75!black}{Option 1 ($P$): ``A zookeeper bottlefeeds a bear
cub.''}
\newline
\textcolor{green!45!black}{Option 2 ($E$): ``A bear cub bottlefeeds a
zookeeper.''}
\newline
\emph{Gold answer}: option 2. \\
\midrule
\begin{tabular}[c]{@{}c@{}}VLind-Bench\\(proposition\\judgment)\end{tabular} &
\includegraphics[width=4.4cm]{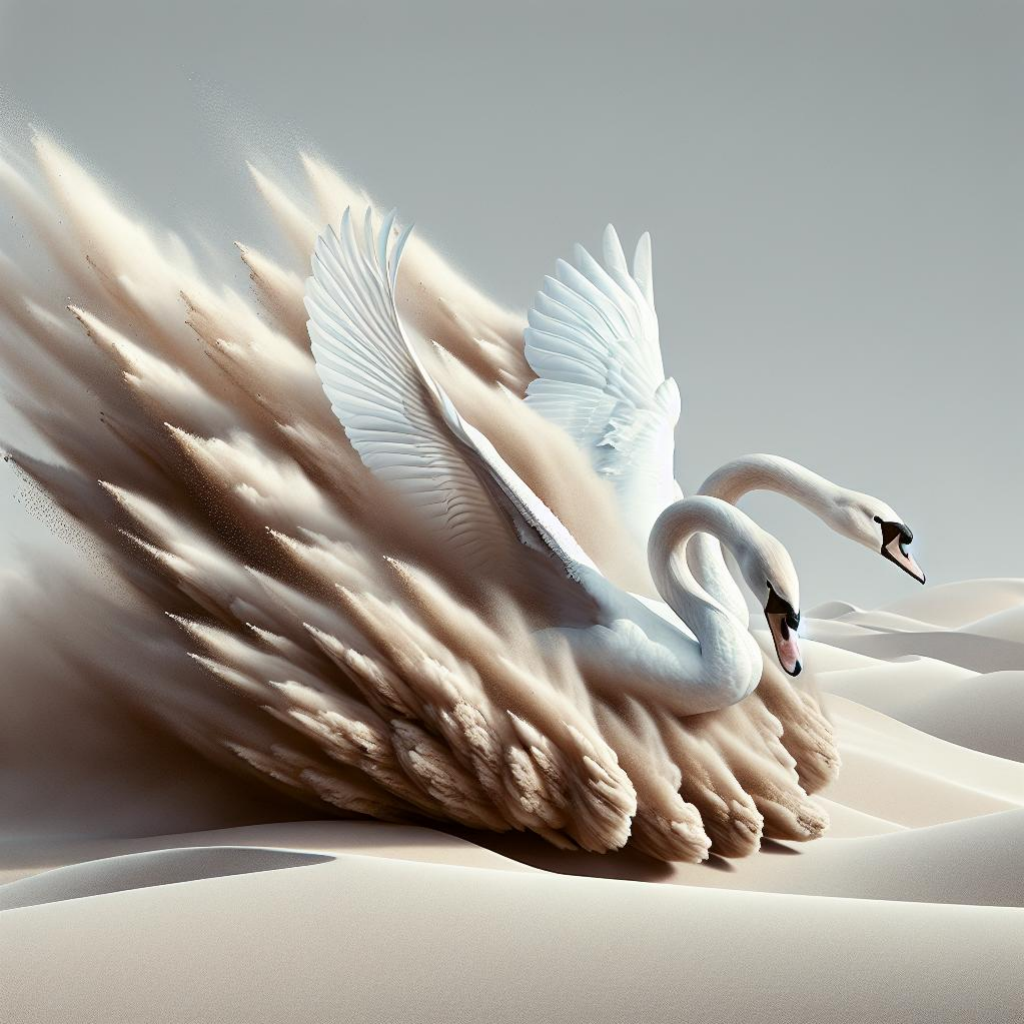} &
\emph{Statement pair, each judged True/False from the image alone}:
\newline
\textcolor{red!75!black}{$P$: ``The swans are aquatic birds found in lakes
and rivers.'' (False given the image)}
\newline
\textcolor{green!45!black}{$E$: ``The swans are found in desert sands.''
(True given the image)} \\
\bottomrule
\end{tabular}
\caption{One real counter-commonsense instance from each evaluation
benchmark. The \textcolor{red!75!black}{red proposition $P$} is what the
language prior favors; the \textcolor{green!45!black}{green proposition $E$}
is what the image actually shows. A prior-driven failure answers $P$
despite the visual evidence for $E$.}
\label{tab:cc-examples}
\end{table*}

\section{Details of the Error-Anatomy Probe}
\label{app:anatomy}
This appendix documents the error-anatomy study summarized in
Section~\ref{sec:prelim-bias}. All runs use the untrained
backbone Qwen3-VL-8B-Instruct served with vLLM, greedy decoding
(temperature $0$), and real CDH-Bench counter-commonsense images. 

\paragraph{Greedy pass and error classification.}
Each of the $285$ counter-commonsense multiple-choice items is asked with the
simple-CoT template ``\texttt{\{question\}}\,/\,\texttt{\{options\}}\,/\,%
\emph{Reason briefly, then end with `Answer: A/B/C/D'}'' (max $400$
tokens). The greedy pass answers $132$ items ($46.3\%$) incorrectly; of
these errors, $109$ ($82.6\%$) select \emph{exactly} the commonsense option
$P$ rather than scattering over the two other distractors, indicating
systematic prior substitution rather than noise.

\paragraph{Grounding probe.}
Every error is re-queried \emph{once} with the same image and options,
temperature $0$ and using the prompt shown in
Table~\ref{tab:anatomy-probe}. This prompt-only change flips $35/132$
errors ($26.5\%$) to the correct answer; $28/132$ ($21.2\%$) are both
prior-biased \emph{and} probe-recovered. We treat the latter as a
conservative lower bound on the ``perception-correct, language-biased''
fraction, since the probe output still passes through the same biased
decoder.

\paragraph{Worked example (over-rationalization).}
Table~\ref{tab:anatomy-probe} also contrasts two trajectories of the
\emph{same} model on the \emph{same} image for the floating-chair item of
Appendix~\ref{app:cc-examples}: the greedy trajectory fabricates
perception to justify $P$, whereas a sampled trajectory faithfully
reports $E$.

\clearpage
\begin{table*}[p]
\centering
\small
\renewcommand{\arraystretch}{1.35}
\begin{tabular}{@{}p{3.4cm} p{12.6cm}@{}}
\toprule
\textbf{Component} & \multicolumn{1}{c@{}}{\textbf{Content}} \\
\midrule
Grounding-probe prompt
&
\itshape Examine the image very carefully. First describe EXACTLY what is
visually present regarding the detail the question asks about ---
count/observe precisely what you SEE, even if it is unusual or contradicts
common sense. Then answer strictly from the image.
\newline
\{question\} \{options\}
\newline
End with `Answer: A', `Answer: B', `Answer: C', or `Answer: D'. \\
\midrule
Greedy trajectory \newline
(\textcolor{red!75!black}{prior-driven, answers $P$})
&
\itshape ``The chair is clearly positioned on the wooden floor, resting on
its legs atop the patterned rug\ldots The visual evidence confirms it is
grounded on the floor. \textcolor{red!75!black}{Answer: A}'' \\
\midrule
Sampled trajectory \newline
(\textcolor{green!45!black}{faithful, answers $E$})
&
\itshape ``The chair is visibly suspended in mid-air, with no visible
contact with the floor or any other surface. Its legs do not touch the rug
or the wooden floor\ldots Therefore, it is floating.
\textcolor{green!45!black}{Answer: C}'' \\
\bottomrule
\end{tabular}
\caption{The grounding-probe prompt and a worked example of
over-rationalization on the floating-chair item of
Appendix~\ref{app:cc-examples}. Given the same image and question, the
greedy trajectory fabricates grounding evidence and returns the
\textcolor{red!75!black}{prior answer $P$ (red)}, while a sampled
trajectory of the same model faithfully reports the
\textcolor{green!45!black}{visual evidence $E$ (green)}.}
\label{tab:anatomy-probe}
\end{table*}


\end{document}